%% file: main.tex
\documentclass[letterpaper]{article} 
\usepackage{aaai24}  
\usepackage{times}  
\usepackage{helvet}  
\usepackage{courier}  
\usepackage[hyphens]{url}  
\usepackage{graphicx} 
\usepackage{natbib}  
\usepackage{caption} 
\usepackage{algorithm}
\usepackage{algorithmic}

\usepackage{subfigure}
\usepackage{booktabs}
\usepackage{multirow}
\usepackage{dsfont}
\usepackage{amsmath}
\usepackage{newfloat}
\usepackage{listings}
\DeclareCaptionStyle{ruled}{labelfont=normalfont,labelsep=colon,strut=off} 
\floatstyle{ruled}
\newfloat{listing}{tb}{lst}{}
\floatname{listing}{Listing}
\title{Measuring Self-Supervised Representation Quality for Downstream Classification using Discriminative Features}
\author {
    Neha Kalibhat\textsuperscript{\rm 1},
    Kanika Narang\textsuperscript{\rm 2},
    Hamed Firooz\textsuperscript{\rm 2},
    Maziar Sanjabi\textsuperscript{\rm 2},
    Soheil Feizi\textsuperscript{\rm 1}
}
\affiliations {
    \textsuperscript{\rm 1}University of Maryland, College Park\\
    \textsuperscript{\rm 2}Meta AI\\
    nehamk@umd.edu
}

\usepackage{bibentry}

\newcommand{\bx}{\mathbf{x}}
\newcommand{\bh}{\mathbf{h}}
\newcommand{\bz}{\mathbf{z}}

\newcommand{\bbR}{\mathds{R}}
\newcommand{\bbone}{\mathds{1}}
\newcommand{\bH}{\mathbf{H}}

\begin{document}

\maketitle



\input{tables_and_figures/disc_feature_to_samples}

\input{abstract}

\input{introduction}

\input{related_work}

\input{tables_and_figures/repsize_vs_accuracy}
\input{ssl_representations}

\input{disc_features}

\input{tables_and_figures/correct_vs_incorrect_latent}
\input{misclassifications}

\input{tables_and_figures/auc_plots}
\input{qscore}

\input{tables_and_figures/acc_baseline_vs_reg}
\input{q_score_results}

\input{tables_and_figures/salient_imagenet_masks}
\input{interpreting_representations}

\input{tables_and_figures/salient_imagenet_iou}
\input{discussion}

\input{acknowledgement}
\bibliography{aaai24}

\newpage
\input{appendix}
\end{document}

%% file: tables_and_figures/disc_feature_to_samples.tex
\begin{figure}
    \centering
    \subfigure{\includegraphics[width = 0.48\textwidth]{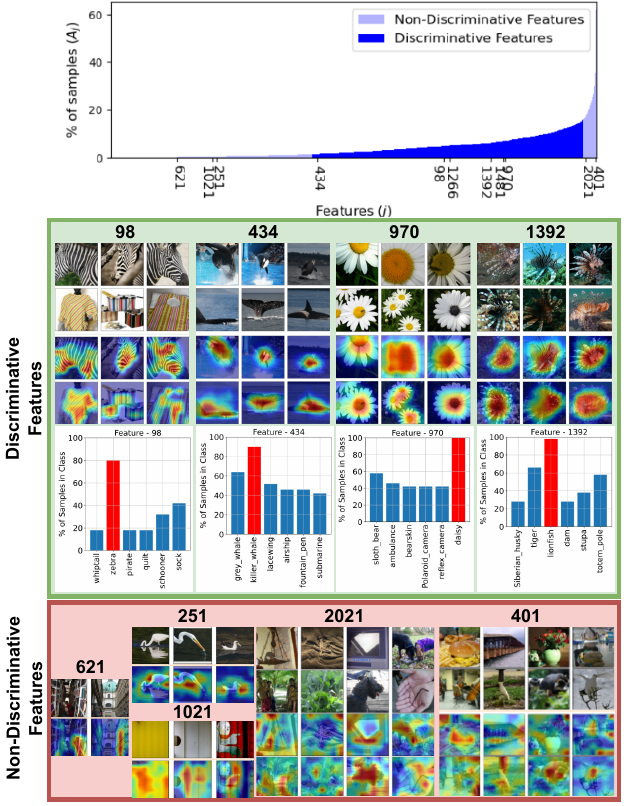}}
    \caption{\textbf{Discriminative Features in Self-Supervised (SSL) Models:} We plot the percentage of highly activating samples for each feature in the SimCLR (ResNet-50) representation space. The features that show very low or very high percentage activations are \textit{non-discriminative} as they likely correspond to very uncommon (lower tail) or very general attributes (upper tail). The features that activate a moderate number of samples (middle portion) are called \textit{discriminative features}. As shown in the gradient heatmaps, these features encode important physical attributes shared among specific classes. These features play a key role in assessing the quality of SSL representations for downstream linear classification tasks.}
    \label{fig:disc_feature_to_samples}
\end{figure}

%% file: abstract.tex
\begin{abstract}

Self-supervised learning (SSL) has shown impressive results in downstream classification tasks. However, there is limited work in understanding their failure modes and interpreting their learned representations. In this paper, we study the representation space of state-of-the-art self-supervised models including SimCLR, SwaV, MoCo, BYOL, DINO, SimSiam, VICReg and Barlow Twins. Without the use of class label information, we discover discriminative features that correspond to unique physical attributes in images, present mostly in correctly-classified representations. Using these features, we can compress the representation space by up to $40\%$ without significantly affecting linear classification performance. We then propose Self-Supervised Representation Quality Score (or Q-Score), an unsupervised score that can reliably predict if a given sample is likely to be mis-classified during linear evaluation, achieving AUPRC of $91.45$ on ImageNet-100 and $78.78$ on ImageNet-1K. Q-Score can also be used as a regularization term on pre-trained encoders to remedy low-quality representations. Fine-tuning with Q-Score regularization can boost the linear probing accuracy of SSL models by up to $5.8\%$ on ImageNet-100 and $3.7\%$ on ImageNet-1K compared to their baselines. Finally, using gradient heatmaps and Salient ImageNet masks, we define a metric to quantify the interpretability of each representation. We show that discriminative features are strongly correlated to core attributes and, enhancing these features through Q-score regularization makes SSL representations more interpretable. 
\end{abstract}

%% file: introduction.tex
\section{Introduction}
Self-supervised models \cite{simclr, swav, mocov2, byol, simsiam, deepcluster, khosla2020supervised, dino, vicreg, barlow} learn to extract useful representations from data without relying on human supervision, and perform comparably to supervised models in downstream classification tasks. Pre-training these models can be highly resource-intensive and time-consuming. It is therefore crucial that the learned representations are of high quality such that they are explainable and generalizable. However, in practice, these representations are often quite noisy and un-interpretable, causing difficulties in understanding and debugging their failure modes \cite{jing2022understanding, huang2021generalization, ericsson2021selfsupervised}.  

In this paper, our goal is to study the representation space of pre-trained self-supervised encoders (SSL) such as SimCLR \cite{simclr}, SwaV \cite{swav}, MoCo \cite{mocov2}, BYOL \cite{byol}, SimSiam \cite{simsiam}, DINO \cite{dino}, VICReg \cite{vicreg} and Barlow Twins \cite{barlow} and discover their informative features in an unsupervised manner. We observe that representations are mostly sparse, containing a small number of \textit{highly activating features}. These features can strongly activate a small, moderate or large number of samples in the population. We refer to the moderate category of features as \textit{discriminative features}. 

We observe some intriguing properties of discriminative features: (i) Although discovered without any class label information, they can be strongly correlated to a particular class or group of classes (See Figure \ref{fig:disc_feature_to_samples}); (ii) They highlight informative concepts in the activating samples which are often related to the ground truth of those samples; (iii) They activate strongly in correctly classified representations rather than mis-classified representations (as shown in Figure \ref{fig:correct_vs_incorrect_latent}) and finally, iv) Representations can be compressed by up to $40\%$ using discriminative features without significantly affecting linear evaluation performance.

Building on these observations, we propose an unsupervised, sample-wise {\bf Self-Supervised Representation Quality Score (Q-Score)}. A high Q-Score for a sample implies that its representation contains highly activating discriminative coordinates which is a favorable representation property. We empirically observe that Q-Score can be used as a zero-shot predictor in distinguishing between correct and incorrect classifications for any SSL model achieving AUPRC of $91.45$ on ImageNet-100 and $78.78$ AUPRC on ImageNet-1K.

We next apply Q-Score as a regularizer and further-train pre-trained SSL models at a low rate to improve low-quality representations. This improves the linear probing performance across all our baselines, highest on BYOL ($5.8\%$ on ImageNet-100 and $3.7\%$ on ImageNet-1K). The representations, after regularization, show increased activation for discriminative features (Figure \ref{fig:correct_vs_incorrect_latent}) due to which several previously mis-classified samples get correctly classified with higher confidence.

Finally, we define a metric for quantifying representation interpretability using Salient ImageNet \cite{singla2021salient} masks as ground truth. Discriminative features are strongly correlated to \textit{core} features of Salient ImageNet. We can potentially explain these features by correlating their meanings with the feature annotations provided for core features in Salient ImageNet. We also observe that discriminative features in mis-classified representations are less correlated with core features compared to correct classifications. Q-score regularization improves this correlation for both correct and mis-classified representations, thereby making representations more explainable. 



%% file: related_work.tex
\section{Related Work}\label{sec:related_work}
Unsupervised methods for classification has been a long-standing area of research, traditionally involving the use of clustering techniques \cite{pmlr-v70-bojanowski17a, NIPS2014_07563a3f, YM.2020Self-labelling, NIPS2016_65fc52ed, Caron_2018_ECCV, Caron_2019_ICCV, pmlr-v97-huang19b}. Self-supervised learning, is a powerful approach that enables learning by preparing own labels for every sample \cite{pmlr-v70-bojanowski17a, NIPS2014_07563a3f, Wu_2018_CVPR, 7312476} usually with the help of a contrastive loss \cite{arora2019theoretical, tosh2021contrastive, NEURIPS2019_ddf35421}. Positive views in SSL losses are multiple transformations \cite{NEURIPS2020_4c2e5eaa} of a given sample using stochastic data augmentation. Through this approach, several state-of-the-art SSL techniques \cite{simclr, swav, simsiam, byol, mocov2, khosla2020supervised} have produced representations that show competitive linear classification accuracy to that of supervised approaches. 

Understanding these learned representations is relatively less explored. Several feature interpretability techniques exist \cite{bau2017network, kalibhat2023identifying,hernandez2022natural}, that aim to explain individual neurons with natural language. However, our goal is to study representations through the lens of failure modes and generalization. \cite{jing2022understanding}, observes that self-supervised representations collapse to a lower dimensional space instead of the entire embedding space. Other methods \cite{Kugelenetal21, xiao2021what}, propose to separate the representation space into variant and invariant information so that augmentations are not task-specific. \cite{grigg2021selfsupervised} observes representations across layers of the encoder and compare it to supervised setups. Clustering-based or prototypical-based methods have also been proposed where the representation space is collapsed into a low-rank space \cite{nnclr, meanshift}. \cite{bordes2021high} uses an RCDM model to understand representation invariance to augmentations. \cite{garrido2022rankme, li2022understanding} propose a score based on the rank of all post-projector embeddings that can be used to judge and compare various self-supervised models. 

In this work, we focus more on studying the properties of representations across correct and incorrect classifications in downstream linear probing (without using any labels). We investigate the connection between these unsupervised properties in the representation space and mis-classifications. Unlike \cite{garrido2022rankme, li2022understanding} which requires computing rank over the entire dataset, our analysis leads to the development of an unsupervised \textit{sample-wise} quality score which can be used as a regularizer and effectively improve downstream classification performance.

%% file: tables_and_figures/repsize_vs_accuracy.tex
\begin{figure}
    \centering
    \subfigure{\includegraphics[width=0.4\textwidth, trim = {0.4cm, 0.5cm, 0.3cm, 0.4cm}, clip]{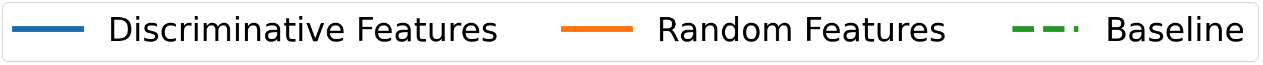}}\\
    \subfigure{\includegraphics[width=0.11\textwidth, trim = {0.4cm, 0.5cm, 0.4cm, 0.5cm}, clip]{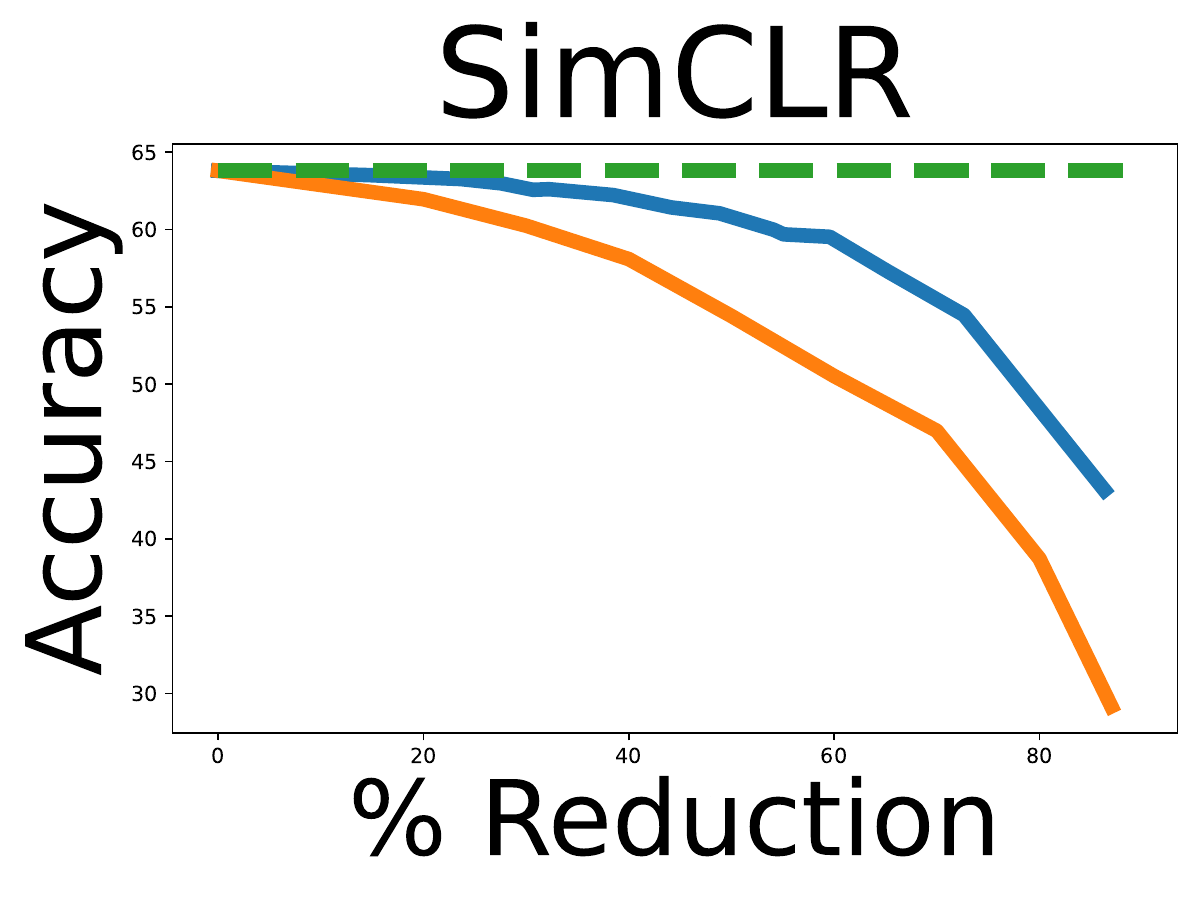}}
    \subfigure{\includegraphics[width=0.11\textwidth, trim = {0.4cm, 0.5cm, 0.4cm, 0.5cm}, clip]{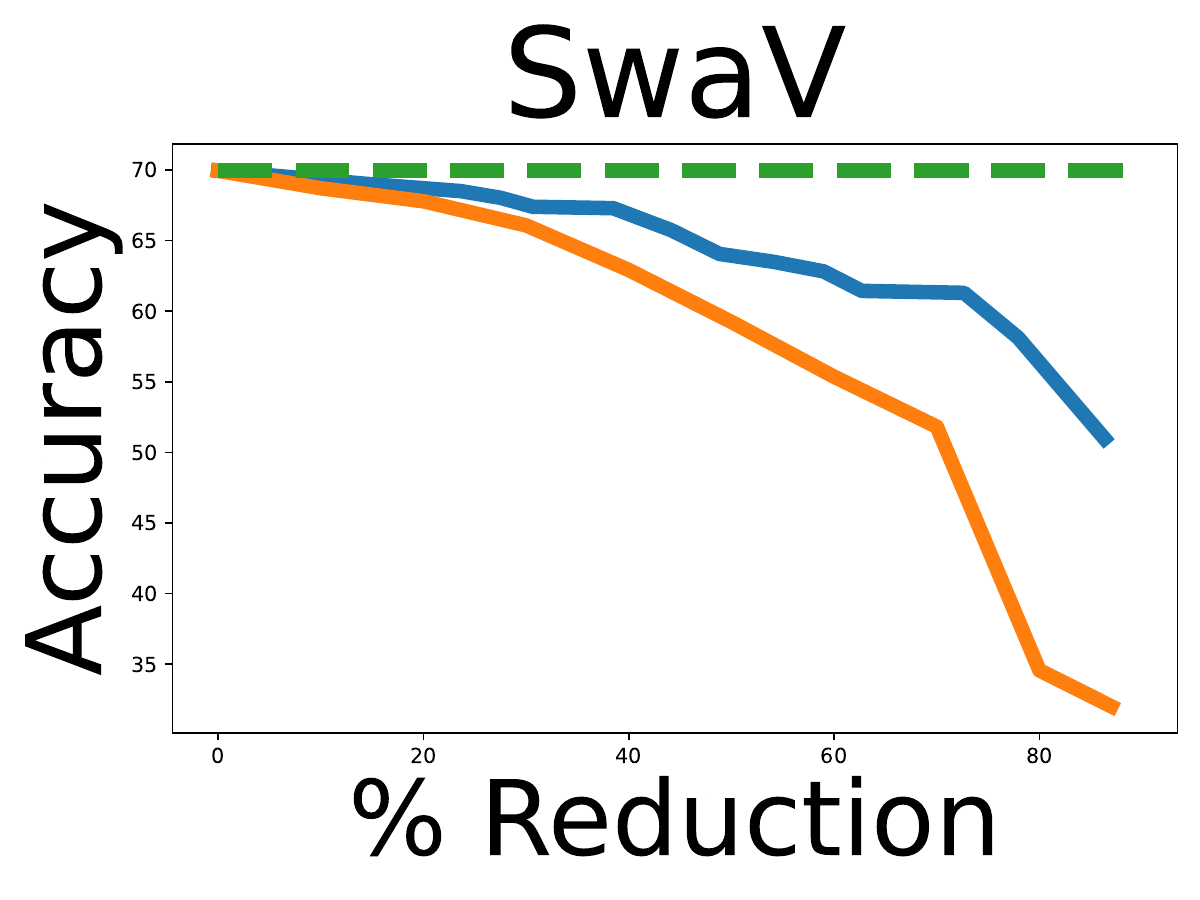}}
    \subfigure{\includegraphics[width=0.11\textwidth, trim = {0.4cm, 0.5cm, 0.4cm, 0.5cm}, clip]{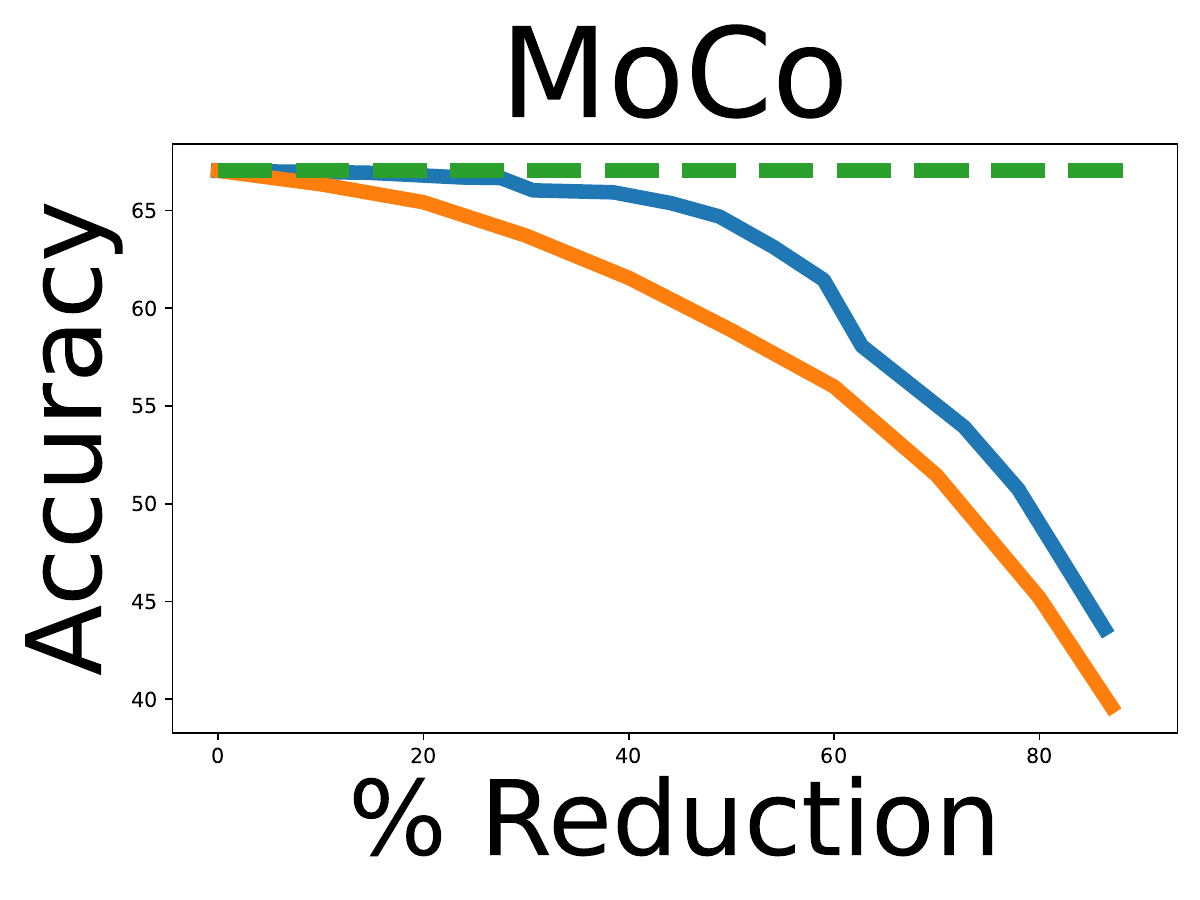}}
    \subfigure{\includegraphics[width=0.11\textwidth, trim = {0.4cm, 0.5cm, 0.4cm, 0.5cm}, clip]{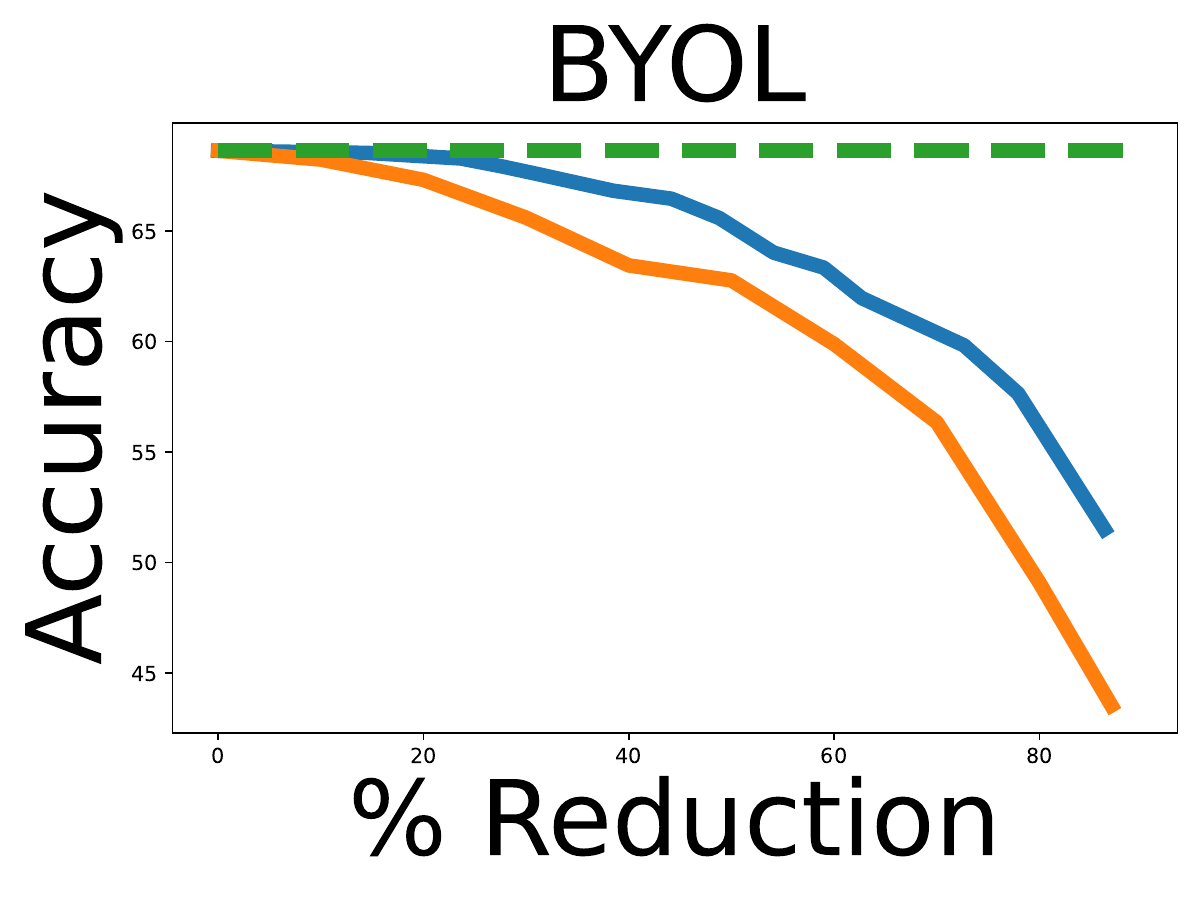}} \\
    \subfigure{\includegraphics[width=0.11\textwidth, trim = {0.4cm, 0.5cm, 0.4cm, 0.5cm}, clip]{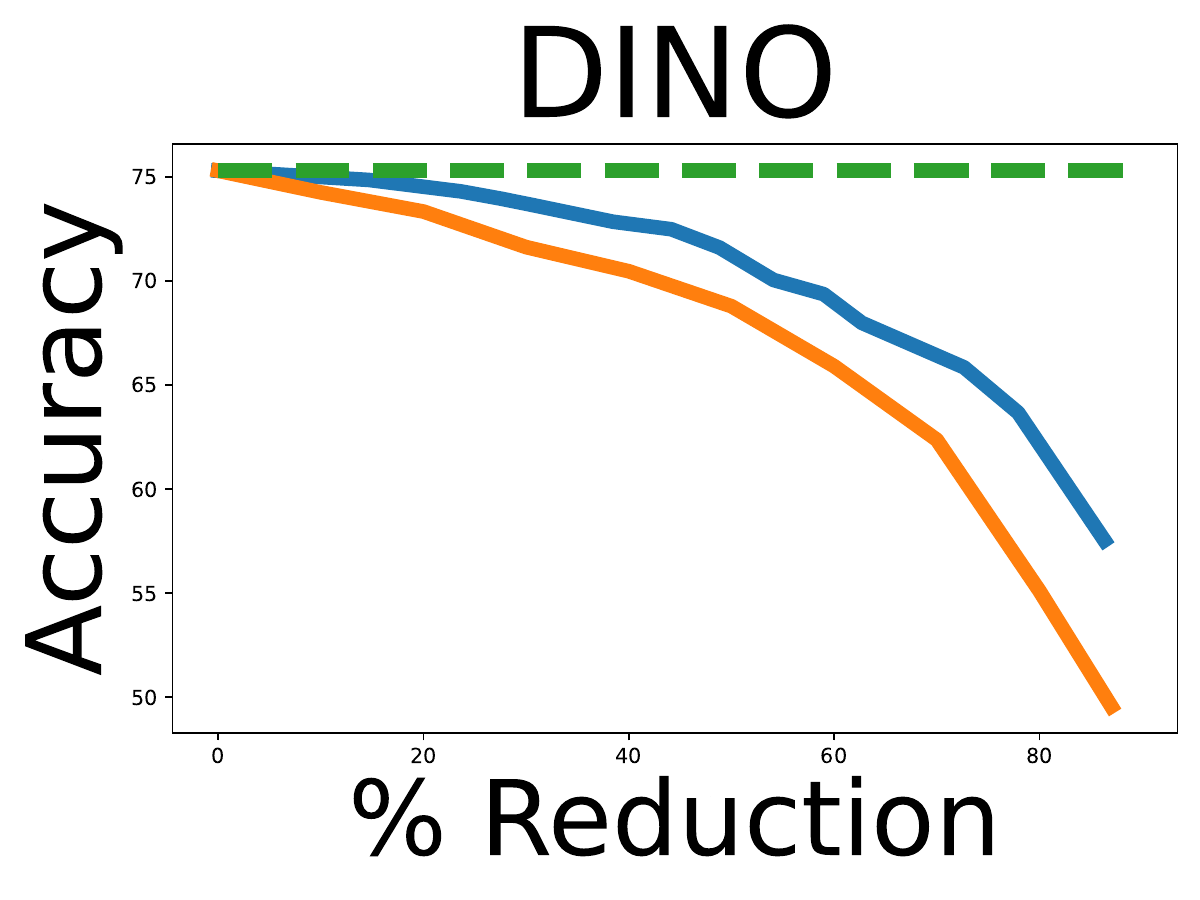}}
    \subfigure{\includegraphics[width=0.11\textwidth, trim = {0.4cm, 0.5cm, 0.4cm, 0.5cm}, clip]{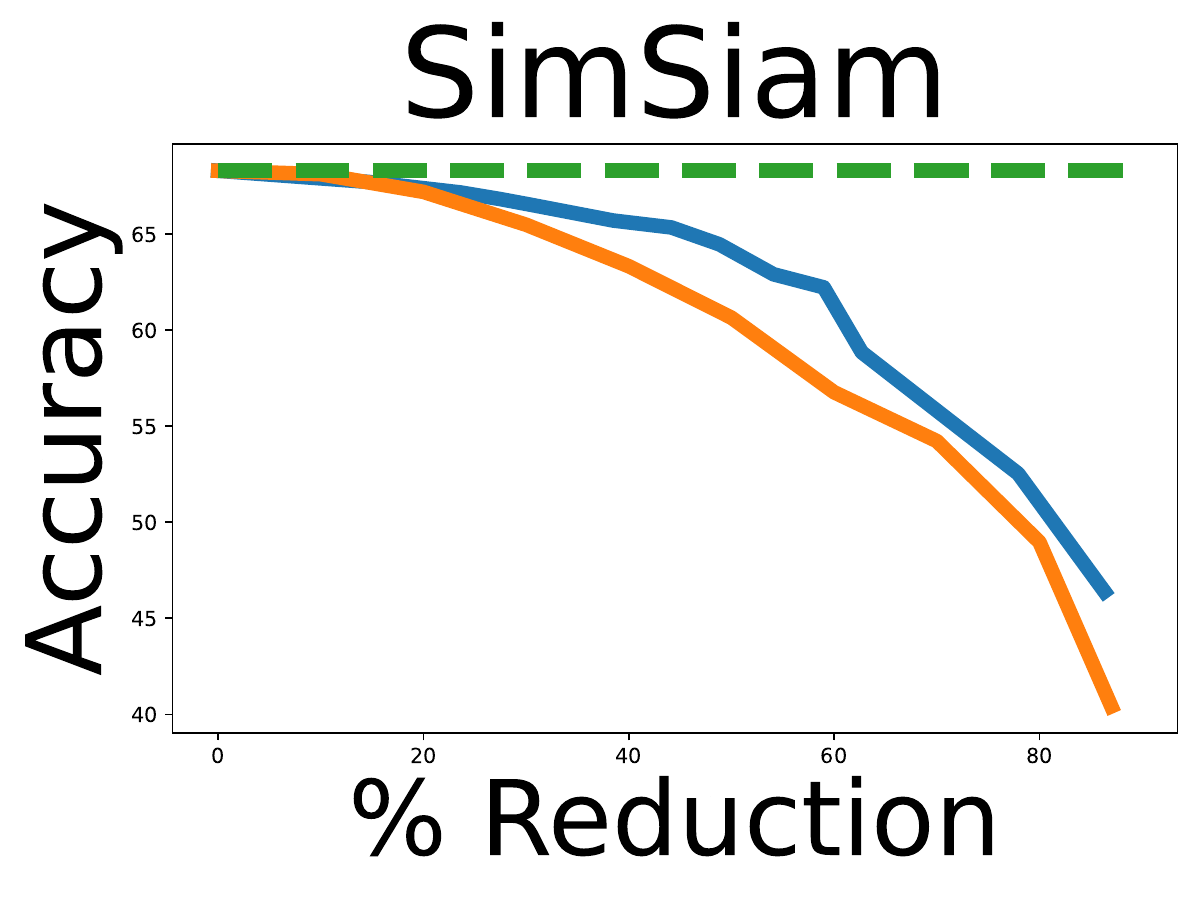}}
    \subfigure{\includegraphics[width=0.11\textwidth, trim = {0.4cm, 0.5cm, 0.4cm, 0.5cm}, clip]{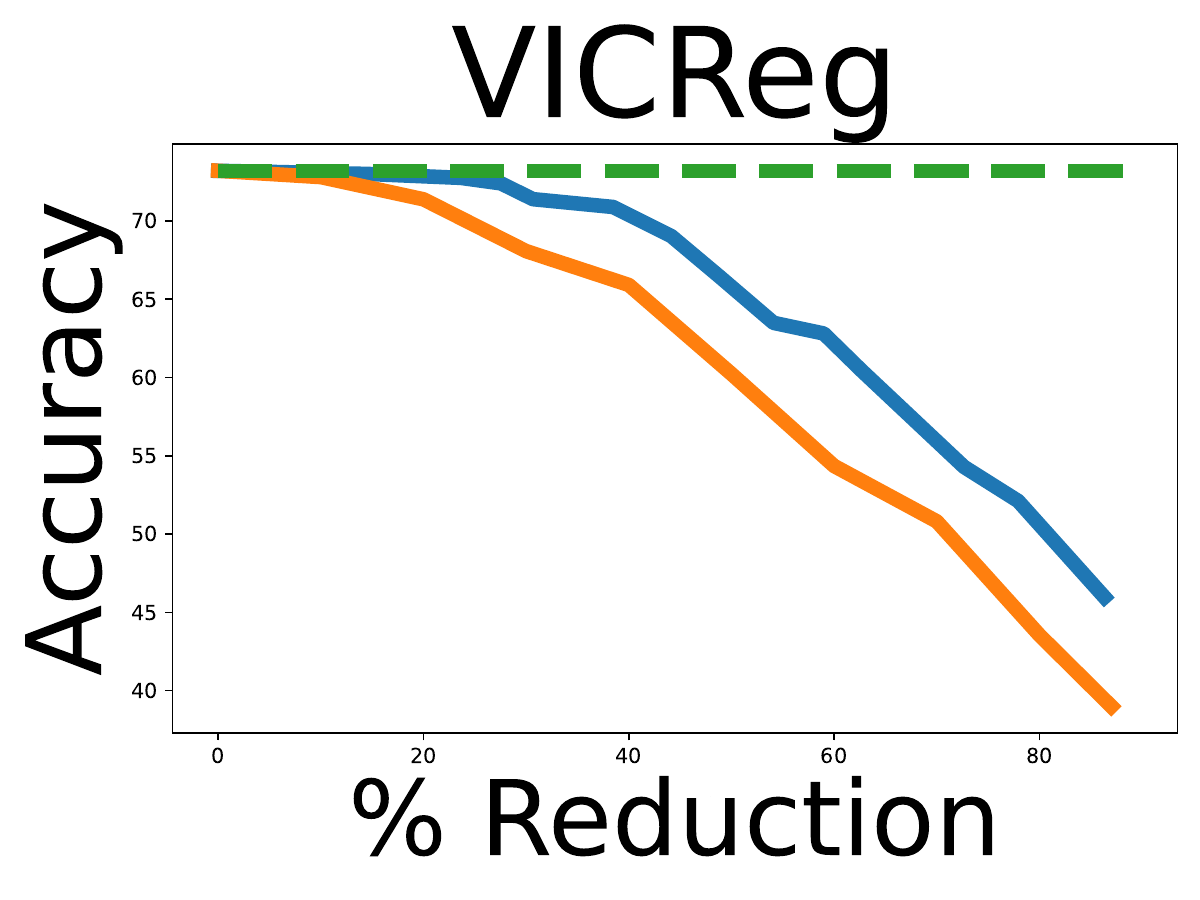}}
    \subfigure{\includegraphics[width=0.11\textwidth, trim = {0.4cm, 0.5cm, 0.4cm, 0.5cm}, clip]{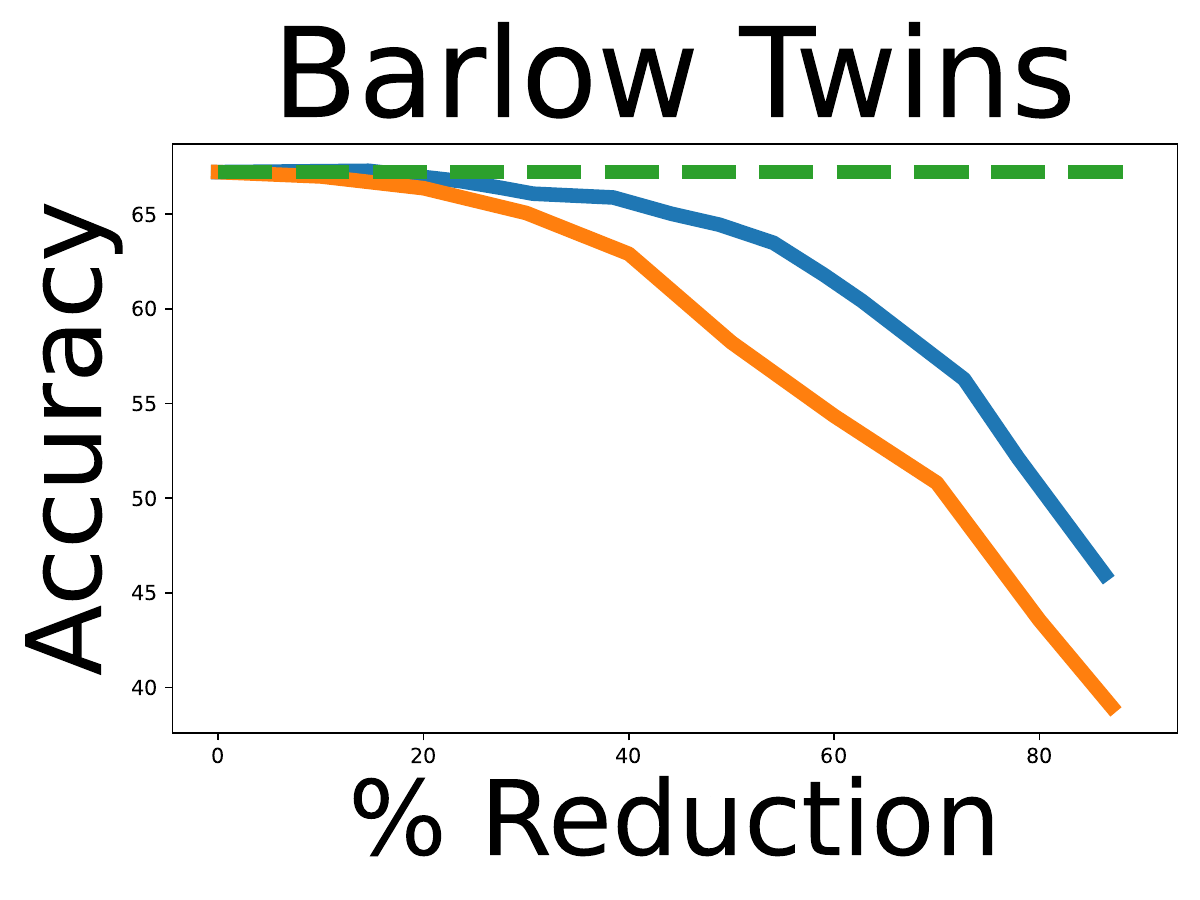}}
    \caption{\textbf{Linear probing discriminative features:} We train linear classifiers after selecting subsets of discriminative features of various sizes (middle portion of Figure \ref{fig:disc_feature_to_samples}) and plot their top-1 accuracy for various SSL baselines. Classifiers trained using discriminative features consistently outperform those of randomly selected features (averaged over 4 random seeds). We can achieve up to $40\%$ reduction in representations size using discriminative features without significantly affecting the top-1 accuracy.}
    \label{fig:repsize_vs_accuracy}
\end{figure}

%% file: ssl_representations.tex
\section{Understanding Representations and their Failure Modes}\label{ssl_representations}
Let us consider a pre-trained self-supervised model with a ResNet \cite{resnet} backbone encoder $f(.)$. Given an input sample, $\bx_i \in \bbR^n$ its representation is denoted by, $f(\bx_i) = \bh_i \in \bbR^r$, where $r$ is the size of the representation space.

Upon visual analysis (See Appendix for more details), we observe that each representation is {\it nearly} sparse, i.e., most feature activations are close to zero \cite{jing2022understanding}. There exists a select few features that are strongly deviated from the remaining features in that representation. For any given representation $\bh_i \in \bbR^r$, we formally define the {\bf set of highly activating features ($L_i$)} as $L_i := \{j : h_{ij} > \mu_i + \epsilon \sigma_i\}$, where $\mu_i$ and $\sigma_i$ denote the mean and standard deviation of $\bh_i$ respectively and $\epsilon$ is a hyperparameter that is empirically selected. We use $\epsilon = 4$ in our experiments. In all our analysis, we perform L2 normalization over every $\bh$ to ensure fair comparison of features. For every feature $j$, the percentage of highly activating samples is denoted by, $A_j = \frac{100}{N} \sum_{i = 1}^{N} \bbone_{j \in L_i}$ where $N$ is the size of the population. In the top panel of Figure \ref{fig:disc_feature_to_samples}, we plot $A_j$ for all features $j$ in the SimCLR representation space of the ImageNet-1K train set. The x-axis is ordered in ascending order of $A_j$. In the next section, we dissect this plot to group features based on $A_j$.

Our observations do not directly extend to ViT-based SSL encoders since, unlike ResNet encoders, their representations can also contain negatively activated features. They need not be sparse in nature (no ReLU before representation layer), rather, features can be both positively (highly activating) or negatively (lowly activating) correlated to important class-specific concepts. In our work, we observe several unique properties of highly activating features in ResNet representations which are beneficial to detect failures. We see ViT encoders as an important direction for future work and focus on ResNet-based encoders for our study.

%% file: disc_features.tex
\subsection{Discriminative Features}
Based on Figure \ref{fig:disc_feature_to_samples}, we can define three broad categories of highly activating features: (i) Features that are highly activating across a very small fraction of the population, corresponding to the lower tail features in Figure \ref{fig:disc_feature_to_samples}. We take the example of features 621, 251 and 1021 and visualize their highly activating samples and gradient heatmaps (using GradCAM \cite{gradcam}). Since these features activate very few samples, they likely correspond to image-specific or uncommon concepts. Such features would also not be useful in classification tasks as these are not shared, class-relevant attributes. (ii) Features that highly activate a very large number of samples in the population i.e, the upper tail features in Figure \ref{fig:disc_feature_to_samples}. Like feature 2021 and 401, such features are likely to encode very broad and general characteristics (like texture, color etc.) common to most samples (spanning various classes) and therefore, are not class-discriminative. The third category includes, (iii) Features that highly activate a moderate number of samples in the population (i.e. the middle part in Figure \ref{fig:disc_feature_to_samples}). These features are most likely to encode unique physical attributes associated with particular classes. For example, feature 98 corresponds to the "stripe" pattern which is an important property of the zebra class. Similarly, feature 970 corresponds to the style of the daisy class, and feature 1392 corresponds to lionfish in different scenes. We refer to this subset of highly activating features as \textit{discriminative features}. Note that we did not use any label information for this analysis. We can identify discriminative and non-discriminative features in a fully unsupervised manner by simply observing their percentage activations ($A$). The bar plots in Figure \ref{fig:disc_feature_to_samples}, show that these features activate more than $80\%$ of particular classes which confirms that these features are strongly class-correlated.

Discriminative features can be regarded as a summarization of the top concepts related to each class of the dataset the encoder is trained on. We justify the described method of selection in Figure \ref{fig:repsize_vs_accuracy}, where we plot the top-1 accuracy of a linear classifier trained on ImageNet-1K using subsets of discriminative features of varying sizes as chosen from Figure \ref{fig:disc_feature_to_samples} (middle portion). We compute the percentile for each point in the distribution $A$ and gradually increase the lower limits (from $0^{th}$ percentile), and decreasing the upper limits (from $100^{th}$ percentile) to get multiple sets of discriminative features of varying sizes. We also plot the top-1 accuracy when random subsets of features are selected. We observe that discriminative features perform significantly better compared to randomly selected features. We also observe that we can reduce the representation size up to $40\%$ using the discriminative features, with minimal reduction in performance. In practice, we select the discriminative features between the $50^{th}$ and the $95^{th}$ percentile of $A$ (as shown in Figure \ref{fig:disc_feature_to_samples}). This range can be discovered empirically and can be further tuned for each model-dataset pair. In the Appendix, we also show that selecting features from either the lower or upper tail of $A$ also under-perform compared to discriminative features from the middle portion.

While we analyze features independently in our work, it has been shown \cite{kalibhat2023identifying, elhage2022toy} that not all neural features are axis-aligned. Meaningful class-related concepts can also be encoded by multiple features (See Appendix for examples). In such cases, the whole \textit{group} of features can be considered as discriminative. These groups can be highly activating for specific classes and lie in the middle portion of $A$. We also perform a PCA analysis on the representation space (see Appendix) to partially validate our selection method. 

%% file: tables_and_figures/correct_vs_incorrect_latent.tex
\begin{figure}
    \centering
    \includegraphics[width=0.48\textwidth]{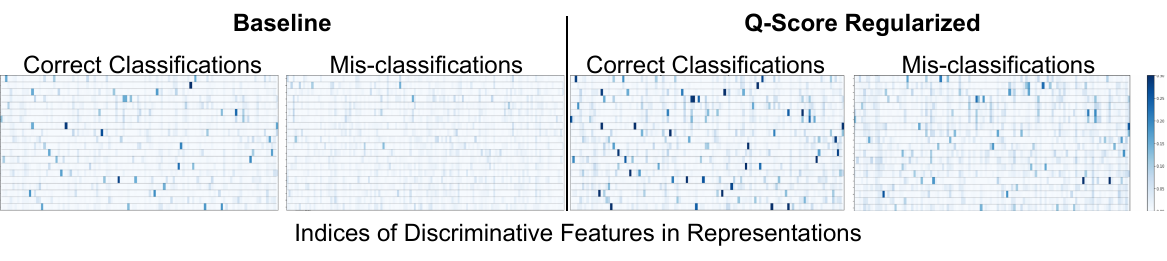}
    \caption{\textbf{Comparing correct and mis-classified representations:} In these heatmaps, we visualize the discriminative features of average SimCLR representations of several ImageNet-1K classes - correct (left) and incorrect (right) classifications. In the baseline, we observe that discriminative features are strongly activated only in correctly classified representations. Q-Score regularization improves discriminative features' activations, even in mis-classified representations.}
    \vspace{-0.2cm}
    \label{fig:correct_vs_incorrect_latent}
\end{figure}

%% file: misclassifications.tex
\subsection{Mis-classified Representations}
We now study how discriminative features play a key role in detecting potential mis-classifications in a fully unsupervised manner. In Figure \ref{fig:correct_vs_incorrect_latent}, we take SimCLR ImageNet-1K representations and visualize the discriminative features. On the left, we show the average representations of correctly classified samples (after linear probing) in a subset of classes, while on the right, we show the same for the mis-classified samples in those classes. The subset of features we display is the same for correct and incorrect classifications.

As we can see, in Figure \ref{fig:correct_vs_incorrect_latent}, in the first panel, there is a clear difference between representations of correctly and incorrectly classified examples. Both correct and mis-classified representations are \textit{nearly} sparse, however, the discriminative features are significantly more activated in correct classifications. This is especially interesting because we can visually distinguish between correct and incorrect classifications, just by observing the discriminative features, without using any label information. Note that this observation does not depend on the actual ground truths or predicted labels of the linear classifier rather, just a binary outcome or whether or not a sample was correctly classified.

The correlation of discriminative features to unique physical attributes as studied in the previous section, suggests that their presence may be useful in correctly classifying representations. In Figure \ref{fig:correct_vs_incorrect_latent}, our claim is confirmed as we observe that mis-classified representations do not show high activations on these features. Therefore, for any given sample, we can consider discriminative features as strong signals indicating classification outcome without requiring to train a linear classification head. We would like to emphasize that our results only indicate an {\it association} between these structural properties and classification accuracy and we do not claim any causal relationship between the two.

%% file: tables_and_figures/auc_plots.tex
\begin{figure}
    \centering
    \subfigure{\includegraphics[width=0.48\textwidth, trim = {0.4cm, 0.6cm, 0.3cm, 0.4cm}, clip]{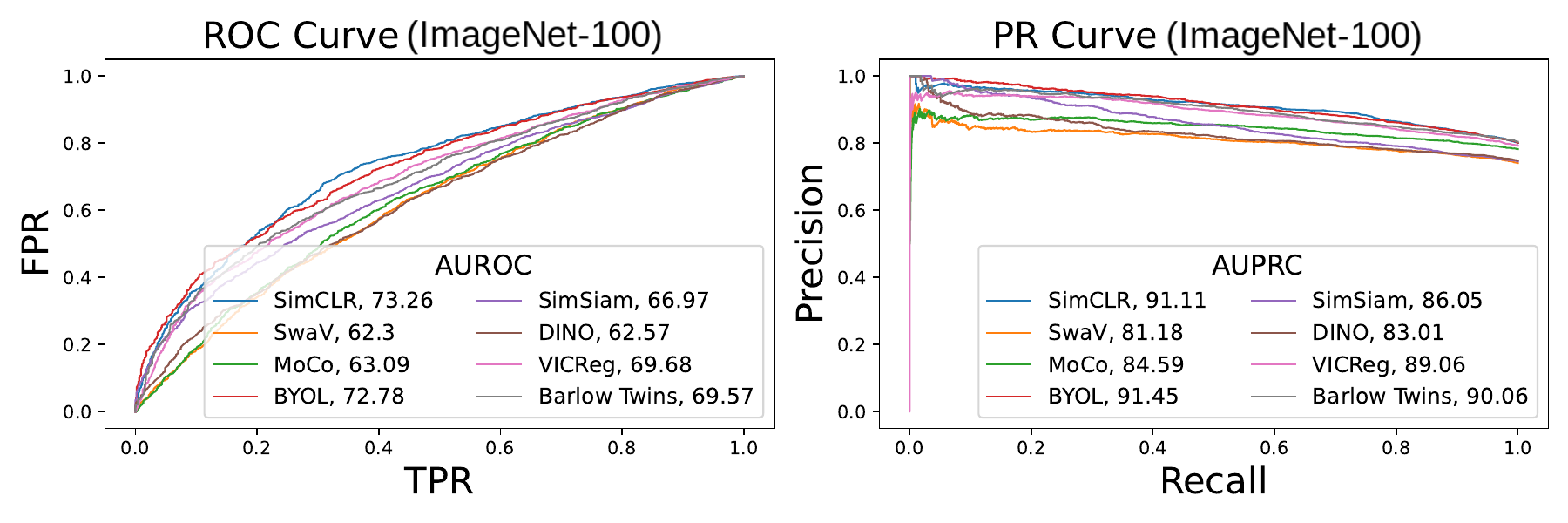}}
     \subfigure{\includegraphics[width=0.48\textwidth, trim = {0.4cm, 0.6cm, 0.3cm, 0.4cm}, clip]{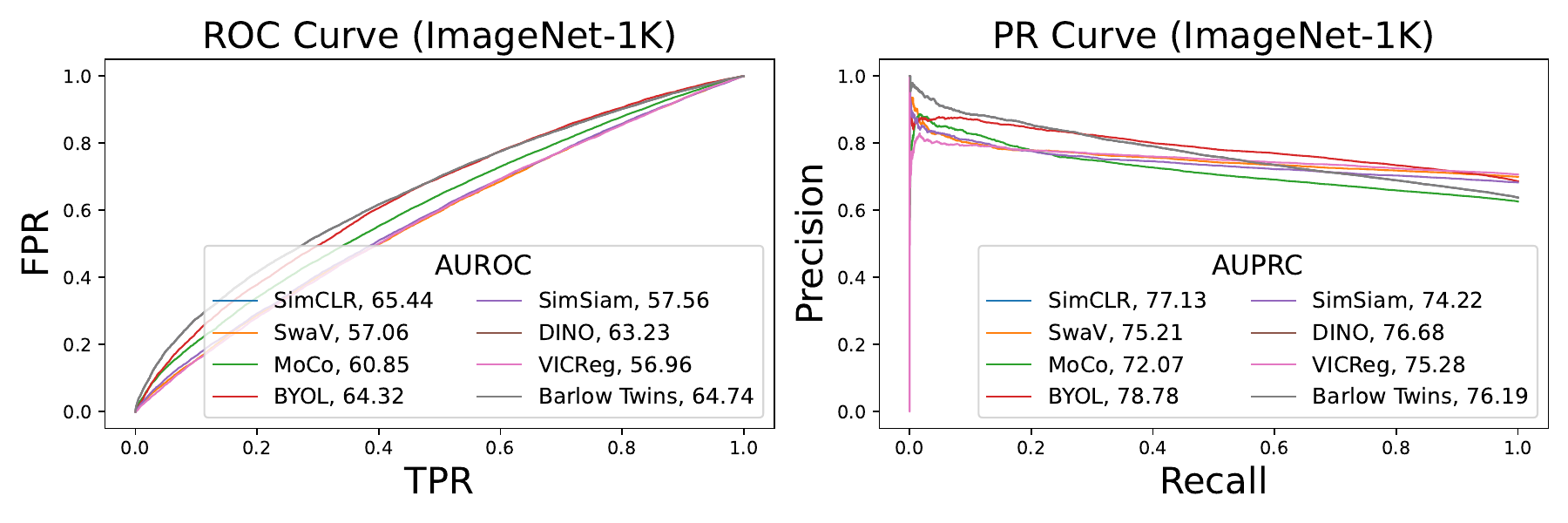}}
    \caption{\textbf{Precision-Recall and ROC curves of  Q-Score:} We measure the effectiveness of Q-Score when used as a predictor in distinguishing between correct and mis-classified representations on ImageNet-100 and ImageNet-1K on each SSL model. Q-Score shows an AUPRC of up to $91.45$ on ImageNet-100, $78.78$ on ImageNet-1K and AUROC of $73.26$ on ImageNet-100, $65.44$ on ImageNet-1K.}
    \label{fig:auc_plots}
\end{figure}

%% file: qscore.tex
\section{Self-Supervised Representation Q-Score}

Our study of learned representation patterns helps us discover discriminative features in an unsupervised manner. These features encode class-specific attributes and help us visually distinguish between correct and incorrect classifications. We combine these observations to design a sample-wise quality score for SSL representations.
Let us define $D$, such that $|D| < r$, as the set of discriminative features for a given SSL model trained on a given dataset. For the $i^{th}$ sample, we have $\bh_i$ (representation), $\mu_i$ (mean of $\bh_i$), $\sigma_i$ (standard deviation of $\bh_i$) and the set of highly activating features $L_i = \{j : h_{ij} > \mu_i + \epsilon \sigma_i\}, |L_i| < r$. We define our Self-Supervised Quality Score for sample $i$ as, 

\begin{align}
    Q_i := \frac{1}{|L_i \cap D|}\sum_{j \in L_i \cap D} (h_{ij} - \mu_i)
\end{align}

where, $L_i \cap D$ is the set of discriminative features specific to the $i^{th}$ sample. Intuitively, higher $Q_i$ implies that the representation contains highly activated discriminative features which are strongly deviated from the mean. Our objective with this metric is to compute a sample-specific score in an unsupervised manner indicating the quality of its representations. Ideally, we would like to argue that samples with higher Q-score have improved representations and thus are more likely to be classified correctly in the downstream task. This is a general score that can be applied to any ResNet-based SSL model trained on any dataset. See Appendix for a discussion on Q-Score in supervised models. 

Next, we measure how effective our score is in differentiating between correctly and incorrectly classified representations in an unsupervised manner. In Figure \ref{fig:auc_plots}, we plot the Precision-Recall (PR) curve and the Receiver Operating Characteristic (ROC) curve of Q-Score when used as a predictor of classification outcome (correct or incorrect). We show this for SimCLR, SwaV, MoCo, BYOL, DINO and SimSiam for ImageNet-100 (top panel) and ImageNet-1K (bottom panel). We also compute the AUROC (area under receiver operating characteristic curve) and AUPRC (area under precision-recall curve) of these curves. We observe AUPRC up to $91.45$ on ImageNet-100 and $78.78$ on ImageNet-1K on BYOL. On SimCLR, we observe AUROC up to $73.26$ on ImageNet-100 and $65.44$ on ImageNet-1K. Based on these results we can conclude that, Q-Score is a reliable metric in assessing the quality of representations, meaning that representations with lower Q-Score (quality), are more likely to be mis-classified.

We now check if promoting Q-Score on pre-trained representations is helpful. To do so, we take state-of-the-art pre-trained SSL models and further train them for a small number of iterations with Q-Score as a regularizer. For example, we can apply this regularizer to the SimCLR optimization as follows,

\begin{align}
    \max_\theta \frac{1}{2N} \sum_{i=1}^{2N}  &\Big[\log \frac{sim(\bz_i, \Tilde{\bz}_i)}{\sum_{j=1}^{2N} \bbone_{j \ne i} sim(\bz_i, \bz_j)} + \lambda_1 \bbone_{Q_i < \alpha}( Q_i) \Big]
\end{align}

where $\bz$ is the latent vector computed by passing $\bh$ through a projector network and $sim(.)$ denotes the exponentiated cosine similarity of the normalized latent vector. $\alpha$ is a threshold with which we select the low-score samples whose Q-Scores should be maximized and $\lambda_1$ is the regularization coefficient. In other words the goal of this regularization is to improve low-quality representations, similar to the ones shown in Figure \ref{fig:correct_vs_incorrect_latent}, by maximizing their discriminative features for downstream classification. 

In practice, directly applying this regularization could lead to a trivial solution where a small set of features get activated for all samples. This is not a favorable situation because these representations become harder to classify accurately and more importantly, the discriminative features are no longer \textit{informative} because they are activated for all samples (similar to the upper tail in Figure \ref{fig:disc_feature_to_samples}). Such features have significantly large L1 norms {\it across} samples compared to the remaining features. Therefore, in our revised optimization, we penalize features that have large L1 norms across samples. Let us denote the representation matrix of a given batch by $\bH \in \bbR^{2N \times r}$ and $\|\bH_{*,k}\|_1$ represents the L1 norm of the $k^{\text{th}}$ column (corresponding to the $k^{\text{th}}$ feature). Our regularized objective would then be, 

\begin{align}
    \max_\theta  \frac{1}{2N} & \sum_{i=1}^{2N} \Big[\log \frac{sim(\bz_i, \Tilde{\bz}_i)}{\sum_{j=1}^{2N} \bbone_{j \ne i} sim(\bz_i, \bz_j)} + \lambda_1 \bbone_{Q_i < \alpha}( Q_i) \Big] \nonumber \\
    &- \lambda_2 \sum_{k = 1}^{r} \bbone_{\|\bH_{*,k}\|_1 > \beta} (\|\bH_{*,k}\|_1)
\end{align}
where the threshold $\beta$ helps us select the uninformative features whose L1 norms should be minimized. In practice, we choose $\alpha$ and $\beta$ for each batch as the mean values of $Q_i$ and $\|\bH_{*,k}\|_1$ respectively.

%% file: tables_and_figures/acc_baseline_vs_reg.tex
\begin{table}
    \centering
    \caption{\textbf{Boosting linear classification performance with Q-Score regularization:} We tabulate the top-1 accuracy of linear evaluation on SimCLR, SwaV, MoCo, BYOL, DINO, SimSiam, VICReg and Barlow Twins with and without Q-Score regularized fine-tuning and a simple lasso regularization. We observe that Q-Score regularization consistently improves each SSL state-of-the-art baseline achieving up to 5.8\% relative improvement on ImageNet-100 and 3.7\% on ImageNet-1K.
    }
    
    \resizebox{0.48\textwidth}{!}{
    
    \begin{tabular}{c|c|c|c|c|c|c}
    \toprule
    \multirow{2}{*}{\textbf{Model}} & \multicolumn{3}{c|}{\textbf{ImageNet-100}} & \multicolumn{3}{c}{\textbf{ImageNet-1K}} \\
    & \textbf{Baseline} & \textbf{Lasso} & \textbf{Q-Score} & \textbf{Baseline} & \textbf{Lasso} & \textbf{Q-Score} \\
    \midrule
    \midrule
    
    SimCLR & 78.64 & 75.63 & \textbf{80.79} (+2.2\%) & 63.80 & 61.48 & \textbf{66.18} (+2.3\%) \\ 
    SwaV & 74.36 & 74.56 & \textbf{78.90} (+4.5\%) & 69.95 & 67.35 & \textbf{71.05} (+1.1\%) \\ 
    MoCo & 79.62 & 78.81 & \textbf{85.16} (+5.5\%) & 67.03 & 65.12 & \textbf{69.31} (+2.2\%) \\ 
    BYOL & 80.88 & 78.73 & \textbf{86.72} (+5.8\%) & 69.14 & 68.47 & \textbf{72.81} (+3.7\%) \\ 
    DINO & 75.41 & 75.18 & \textbf{76.39} (+1.0\%) & 75.52 & 72.89 & \textbf{75.78} (+0.3\%) \\ 
    SimSiam  & 78.80 & 78.42 & \textbf{81.41} (+2.6\%) & 68.62 & 68.63 & \textbf{70.47} (+1.9\%) \\ 
    VICReg & 79.77 & 76.95 & \textbf{81.56} (+1.8\%) & 73.63 & 72.86 & \textbf{74.72} (+1.1\%) \\ 
    Barlow & 80.63 & 80.32 & \textbf{81.03} (+0.4\%) & 67.85 & 66.47 & \textbf{69.58} (+1.7\%) \\ 
    \bottomrule
    \end{tabular}
    
    }
    \label{tab:acc_baseline_vs_reg}
\end{table}

%% file: q_score_results.tex
\subsection{Experimental Setup}
Our setup consists of state-of-the-art self-supervised ResNet encoders ($f(.)$) - SimCLR, SwaV, MoCo, BYOL, DINO (ResNet-based), SimSiam, VICReg and Barlow Twins that are pre-trained on datasets - ImageNet-1K, ImageNet-100 \cite{imagenet}. We use a ResNet-50 encoder for our ImageNet-1K experiments and ResNet-18 encoder for all other datasets. We discover discriminative features for each pre-trained model using the train set of each dataset. For Q-Score regularization, maintaining the same encoder architecture as the respective papers, we use the LARS \cite{You2017LargeBT} optimizer with warmup-anneal scheduling. We further-train each pre-trained model with and without Q-Score regularization (controlled by $\lambda_1$ and $\lambda_2$) using a low learning rate of $10^{-5}$ for $50$ epochs. We find that $\lambda_1 = \lambda_2 = 10^{-4}$ generally works well for fine tuning. We use a maximum of 4 NVIDIA RTX A4000 GPUs (16GB memory) for all our experiments. Using the implementations from solo-learn \cite{solo}, we have tried to match our baseline numbers as much as possible within the error bars reported in the papers using the available resources. We follow the standard evaluation by training a linear classifier on frozen pre-trained representations for $100$ epochs. For all our gradient heatmap visualizations, we utilize GradCAM \cite{gradcam}.

\subsection{Q-Score Regularization}
We tabulate our linear evaluation results of various SSL baselines before and after Q-Score regularization in Table \ref{tab:acc_baseline_vs_reg}. We also include results on lasso (L1) regularization \cite{tibshirani1996regression} on pre-trained models. Lasso promotes sparsity by minimizing the L1 norm of representations.  Q-Score regularization improves the linear probing top-1 accuracy on all of the SSL state-of-the-art models. We observe the most improvement on BYOL showing $5.8\%$ increase in accuracy on ImageNet-100 and $3.7\%$ on ImageNet-1K. Lasso regularization shows degraded performance across most models since naively sparsifying representations can lead to loss of information. In contrast, Q-Score regularization promotes highly activating discriminative coordinates which we have shown to be essential for downstream classification. We include more results on CIFAR-10 \cite{cifar10}, STL-10 \cite{stl10} and CIFAR-100 \cite{cifar100} in the Appendix. We also include the results on the transfer performance of discriminative features and Q-Score regularized ImageNet-1K models on unseen datasets in the Appendix. Q-Score is therefore a powerful regularizer that can boost the performance of state-of-the-art SSL baselines. 

In addition to top-1 accuracy, Q-Score also shows significant improvement in representation quality. In Figure \ref{fig:correct_vs_incorrect_latent}, we compare the discriminative features of representations before and after Q-Score regularization. We observe that the magnitude of discriminative features and consequently the Q-Score, increases for both correct and mis-classified representations after regularization, making it harder to differentiate between them. For example on SimCLR ImageNet-1K, the AUROC reduces to $59.81$ and AUPRC to $71.28$. We also observe improved classification confidence as representations become more disentangled (see Appendix for a discussion on this). Our regularization produces better quality representations with clear discriminative features making them more distinguishable across classes and therefore, easier to classify. Due to this, we can attribute the improvement in performance to improved representation quality. Although Q-Score improves accuracy, it does not entirely prevent mis-classifications as mis-classifications may occur due to a variety of reasons such as, training augmentations, hardness of samples, encoder complexity, dataset imbalance etc. 

\input{tables_and_figures/grad_after_reg}

Our motivation for using discriminative features as discussed in Section \ref{ssl_representations} is because - a) they are at clear contrast between correct and incorrect classifications, and b) they show strong correlation to ground truth. We observed in Figure \ref{fig:correct_vs_incorrect_latent} in the baseline, that the discriminative features in correctly classified samples are not strongly activated in mis-classified samples. We now study some mis-classified samples and observe how their features may improve with Q-Score regularization. In Figure \ref{fig:grad_after_reg}, we visualize the gradient heatmaps of the discriminative features of some mis-classified examples in SimCLR. In the baseline, we observe that discriminative features do highlight portions of the image relevant to the ground truth, however, they may also activate other portions that are not necessarily important (see rock crab and green mamba). These heatmaps reflect low quality representations where the discriminative features are not strongly deviated from the mean. After Q-Score regularization, the maximization of discriminative features also leads to better gradient heatmaps that are more localized and cover almost all important portions of the image relevant to the ground truth. Therefore, these samples get classified correctly with higher confidence after regularization.

%% file: tables_and_figures/grad_after_reg.tex
\begin{figure}
    \centering
    \includegraphics[width = 0.48\textwidth, trim = {0.0cm, 0.2cm, 0.2cm, 0.1cm}, clip]{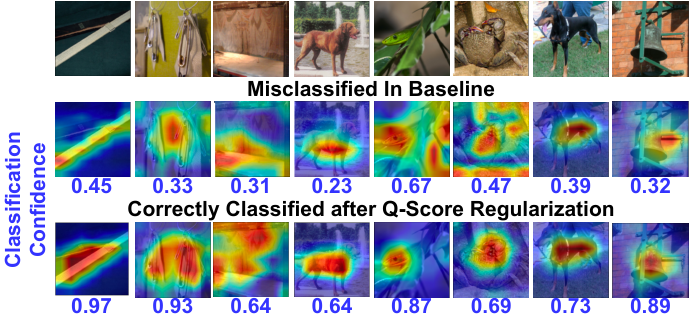}
    
    \caption{\textbf{Discriminative features in mis-classified samples:} The discriminative features' heatmaps on the SimCLR (baseline) activate portions that may not be relevant to the image ground truth, leading to incorrect predictions. After Q-Score regularization on these representations, the heatmaps become more localized and less noisy, whilst improving predictions and confidence.}
    \label{fig:grad_after_reg}
\end{figure}

%% file: tables_and_figures/salient_imagenet_masks.tex
\begin{figure}[h]
    \centering\
    \includegraphics[width=0.48\textwidth,trim={0.0cm, 0.3cm, 0.1cm, 0.1cm},clip]{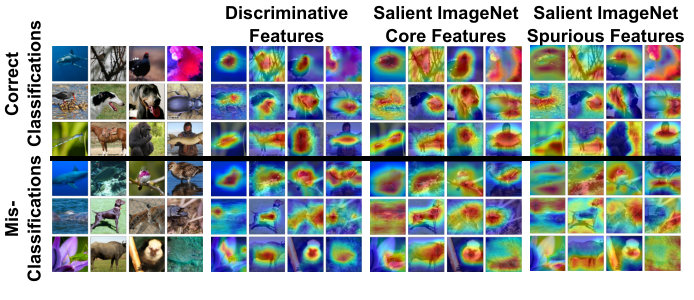}

    \caption{\textbf{Comparing discriminative features with Salient ImageNet \textit{core} and \textit{spurious} features:} We compare the gradient heatmaps of discriminative features correct and incorrect classifications of SimCLR on ImageNet-1K with the core and spurious masks of the same images in Salient ImageNet. We observe that discriminative features generally overlap more with core features in Salient ImageNet. 
    }
    \label{fig:salient_imagenet_masks}
\end{figure}

%% file: interpreting_representations.tex
\section{Quantifying Representation Interpretability with Salient ImageNet}
We have observed that discriminative features in representations correspond to meaningful physical attributes through gradient heatmaps and they play a key role in deciding the downstream classification outcome. In this section, we quantify the interpretability of these features between correct and incorrect classifications. We utilize Salient ImageNet \cite{singla2021salient} as the ground truth baseline to compare our gradient heatmaps with. The Salient ImageNet dataset contains annotated masks for both \textit{core} and \textit{spurious} features extracted from a supervised robust ResNet-50 model for $6858$ images spanning $327$ ImageNet classes. It also contains some natural language keywords, provided by workers to explain each feature. Core features are those that are highly correlated with the ground truth of the image, whereas, spurious features are those that activate portions irrelevant to the ground truth. In Figure \ref{fig:salient_imagenet_masks}, we study some correct and mis-classified samples in the SimCLR baseline. We plot the gradient heatmaps of the discriminative features (combining each individual feature heatmap) of SimCLR for each respective image. We also plot the core and spurious masks of the same images from the Salient ImageNet dataset. We observe that discriminative SimCLR features mostly capture relevant and defining characteristics of the images, therefore are highly correlated with the ground-truth. Moreover, for every correctly classified image, these heatmaps overlap more with core features than spurious features in Salient ImageNet. Discriminative features in mis-classified images also overlap with core features in most cases. Since discriminative features are very closely related (in terms of overlap) to \textit{core} features, we can potentially explain these features better with the help of worker annotations in Salient ImageNet. Therefore, these features can be considered as \textit{interpretable}.

We quantitatively measure the interpretability of a given representation of a given model by computing the Intersection over Union (mIoU) between the heatmap of discriminative features and the core or spurious mask of that image in Salient ImageNet. We can extend this to measure the overall interpretability of a given model by computing the mean Intersection over Union (mIoU) over the population. For the $i^{th}$ image , we define $Ar_i$ as the area of the heatmap of its discriminative features. Let $Ar^{core}_i$ and $Ar^{sp}_i$ be the area of the core and spurious masks respectively. The mIoU scores are defined as follows,

\begin{align*}
    \text{mIoU}^{core} &= \frac{1}{N} \sum_i \frac{s(Ar_i \cap Ar^{core}_i)}{s(Ar_i \cup Ar^{core}_i)} \\
    \text{mIoU}^{sp} &= \frac{1}{N} \sum_i \frac{s(Ar_i \cap Ar^{sp}_i)}{s(Ar_i \cup Ar^{sp}_i)}
\end{align*}

\noindent where $s(.)$ calculates the sum of the pixel values of the discriminative features' heatmap in the given area. Higher $\text{mIoU}^{core} \%$ indicates that, on an average higher percentage of the feature heatmap overlaps with the annotated core region, meaning that the model features are more interpretable. 

In Figure \ref{fig:salient_imagenet_iou}, we show that for all SSL baselines, $\text{mIoU}^{core} > \text{mIoU}^{sp}$ for both correct and incorrect classifications which confirms that discriminative features generally encode important and core attributes over the whole population. Among correct and mis-classified samples in the baselines, we observe that the $\text{mIoU}^{core}$ of correct classifications is higher than mis-classifications. This aligns with our observations in Figure \ref{fig:grad_after_reg}, which shows that discriminative features in mis-classified samples may not be strongly deviated from the mean and therefore, may correspond to less important portions of the image. After Q-Score regularization, we observe an increase in $\text{mIoU}^{core}$ for both correct and mis-classified samples compared to the baseline. This shows that our regularization which enhances discriminative features produces better gradient heatmaps which are more overlapped with core portions of images and therefore, improves the overall model interpretability. Note that, spurious feature heatmaps (Figure \ref{fig:salient_imagenet_masks}) cover almost all the image content. As shown in Figure \ref{fig:grad_after_reg}, discriminative feature heatmaps, after regularization, become larger or smaller to capture all core characteristics of the image. This can cause $\text{mIoU}^{sp}$ to be higher or lower after regularization. We therefore only use $\text{mIoU}^{sp}$ to compare with $\text{mIoU}^{core}$ and not to analyze the regularization effect.

%% file: tables_and_figures/salient_imagenet_iou.tex
\begin{figure}
    \centering
    \subfigure{\includegraphics[width=0.23\textwidth, trim = {0.4cm, 0.6cm, 0.3cm, 0.4cm}, clip]{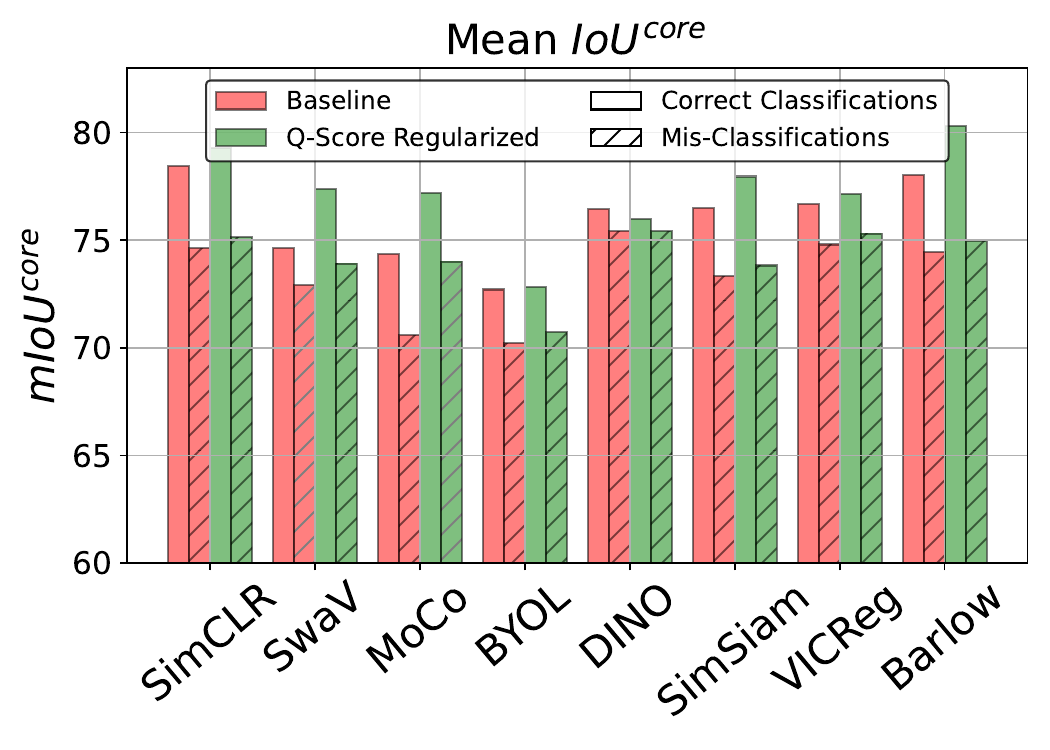}}
    \subfigure{\includegraphics[width=0.23\textwidth, trim = {0.4cm, 0.6cm, 0.3cm, 0.4cm}, clip]{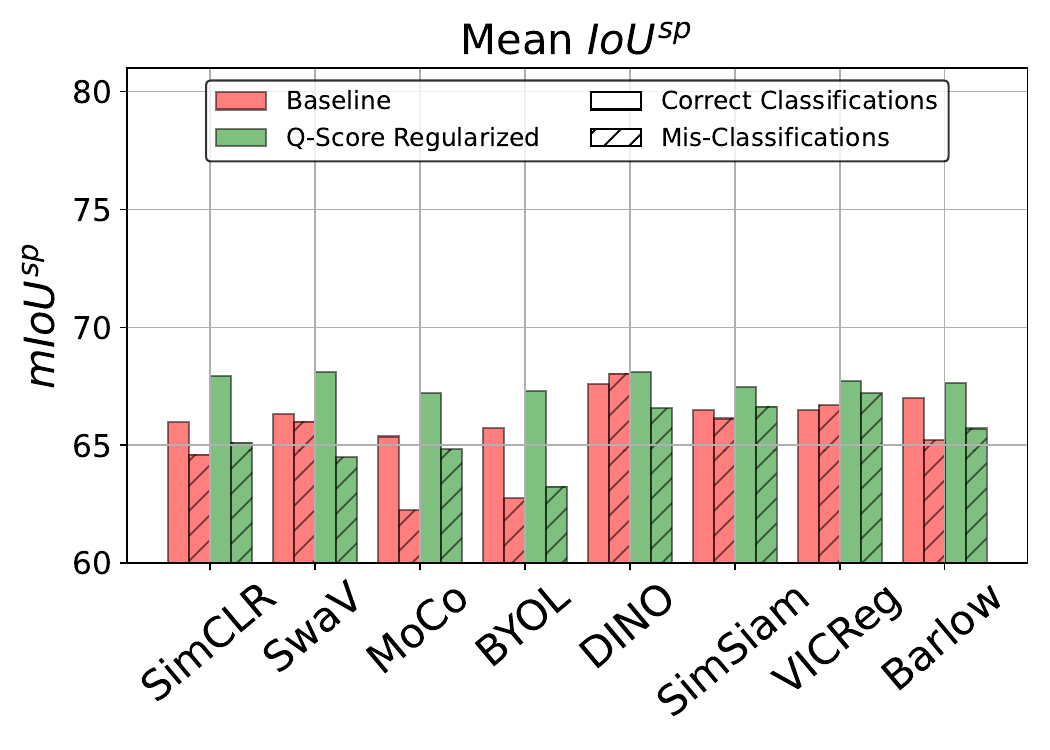}}
    \caption{\textbf{mIoU scores with Salient ImageNet features:} We compute the mean $\text{mIoU}^{core}$ and $\text{mIoU}^{sp}$ scores of SSL baselines (using their discriminative features) before and after Q-Score regularization. We observe that discriminative features for all models generally show higher \% IoU with core features than spurious features. Mis-classified representations show relatively lower \% IoU with core features. After Q-score regularization, we observe that $\text{mIoU}^{core}$ generally improves for both correct and mis-classified representations.}
    \label{fig:salient_imagenet_iou}
\end{figure}

%% file: discussion.tex
\section{Conclusion}
We studied the representation space of SSL models to identify discriminative features in a fully unsupervised manner. Using discriminative features, we compress the representation space by up to $40\%$ without largely affecting the downstream performance. We defined an unsupervised sample-wise score, Q-Score, that uses discriminative features and is effective in determining how likely samples are to be correctly or incorrectly classified. We regularized with Q-Score and remedied low-quality samples, thereby, improving the overall accuracy of state-of-the-art SSL models on ImageNet-1K by up to $3.7\%$, also producing more explainable representations. Our work poses several questions and directions for future work: (i) What are other causes for failures and poor generalization in SSL models apart from representation quality?; (ii) Studying properties for ViT-based representations in a similar fashion is a crucial next step to our work as ViT is a widely used SSL encoder; (iii) How can representations be better structured for non-classification-based downstream tasks such as object detection and semantic segmentation, where discriminative features should correlate to the object/segment categories (which can be several per-image). 



%% file: acknowledgement.tex
\section{Acknowledgement}
This project was supported in part by a grant from an NSF CAREER AWARD 1942230, ONR YIP award N00014-22-1-2271, ARO’s Early Career Program Award 310902-00001, Meta grant 23010098, HR00112090132 (DARPA/RED), HR001119S0026 (DARPA/GARD), Army Grant No. W911NF2120076, NIST 60NANB20D134, the NSF award CCF2212458, an Amazon Research Award and an award from Capital One.

%% file: appendix.tex
\appendix
\onecolumn
\renewcommand\thefigure{A.\arabic{figure}}
\setcounter{figure}{0} 

\renewcommand\thetable{A.\arabic{table}}
\setcounter{table}{0} 

\section{Appendix}

\subsection{Representation Space}
In the top panel of Figure \ref{fig:representation_space}, we visualize the representations of SimCLR pre-trained on ImageNet-1K \cite{imagenet}. Each row denotes the representation vector ($\bh_i$) of a random sample drawn from the the ImageNet-1K train set. There are $2048$ columns corresponding to the representation size of a ResNet-50 \cite{resnet} encoder. 

We observe that the representations are all nearly sparse with a small number of strongly deviated coordinates. We verify this observation in the second panel of Figure \ref{fig:representation_space} where, we plot the distribution of all the SimCLR features of the same samples as the top panel, as well that of other self-supervised models including, DINO, SwaV, MoCo, VICReg and Barlow Twins. In each distribution, a very large number of features have a magnitude of 0 or very close to 0. In the zoomed version of the same plot, we can see a relatively small number of features that show strong activations.

\input{tables_and_figures/representation_space}

\subsection{Results on Other Datasets}
As an extension to the results shown in Table \ref{tab:acc_baseline_vs_reg}, we include results on more datasets including CIFAR-10 \cite{cifar10}, CIFAR-100 \cite{cifar100} and STL-10 \cite{stl10} on 8 self-supervised baselines when fine-tuned (further trained) with and without Q-Score regularization. In Table \ref{tab:acc_other_datasets}, we observe that Q-Score regularization helps boost the performance of all state-of-the-art models across datasets. 

\input{tables_and_figures/results_non_imagenet}

\subsection{Transfer Performance of Q-Score Regularization}
In Table \ref{tab:acc_other_datasets}, we tabulate the transfer learning performance (linear evaluation) of various unseen datasets \cite{cifar10, cifar100, stl10, aircraft, flowers, food, cars, dtd} on 6 self-supervised models trained on ImageNet-1K with and without Q-Score regularization. We use frozen ResNet-50 representations for each transfer dataset (using actual image size) and perform linear evaluation using a classifier. We observe that the average accuracy of unseen datasets improves on all setups, especially on SimCLR, SwaV and MoCo. 

In Figure \ref{fig:grad_transfer}, we visualize the gradient heatmaps of some discriminative features discovered on SimCLR on ImageNet-1K on both ImageNet-1K and unseen datasets, Aircraft \cite{aircraft}, Food \cite{food} and Cars \cite{cars}. We observe that the physical meaning associated with each feature is consistent between both the training and unseen data. The heatmaps also correspond to informative features, strongly correlated with the ground truth. These gradients indicate that discriminative features are transferable across unseen datasets, which support the improvement we observe in Table \ref{tab:acc_other_datasets}. 

We also visualize the representations of correct and incorrect classifications of the Flowers \cite{flowers} dataset in Figure \ref{fig:correct_vs_incorrect_flowers}. We use SimCLR pre-trained on ImageNet-1K (top panel) and the same model pre-trained with Q-Score regularization (bottom panel). We observe that the same properties as Figure \ref{fig:correct_vs_incorrect_latent} on ImageNet-1K (train dataset) transfer at test time to Flowers, an unseen dataset. Before regularization, representations, especially the mis-classified ones, do not contain highly activating discriminative features. These features get more enhanced after Q-Score regularization leading to improved top-1 accuracy as shown in Table \ref{tab:acc_other_datasets}.

\input{tables_and_figures/other_datasets}
\input{tables_and_figures/grad_transfer}
\input{tables_and_figures/correct_vs_incorrect_flowers}

\subsection{Axis-Alignment and Principal Components}
In our analysis, we select discriminative features independently and observe their heatmaps and activations across the population. Figure \ref{fig:non_axis_align} shows some examples of non-axis-aligned groups of features that can still correspond to meaningful concepts associated with class labels. With discriminative features, we attempt to collect all the conceptual information associated with class labels in the dataset. These concepts can be encoded by independent or groups of features and which strongly activate when the concept is present and can still lie in the middle portion of $A$. 

To (partially) validate our selection method, we have also conducted a PCA analysis where we select principal components of feature representations and perform linear evaluation on top of them. In Figure \ref{fig:repsize_vs_accuracy}, we observe that, until $40\%$ reduction of the representation size, PCA and discriminative features perform comparably in terms of the linear classification accuracy while discriminative features significantly outperforms random features across the board. We also plot the gradients of the highly activating PCA features and compare them to discriminative features in the full representation space in Figure \ref{fig:imagenet_pca_grad}. We observe that both sets of features activate the same portions of the images between both correct and incorrect classifications. These results indicate that discriminative features capture a fair amount of information in the feature representations and thus (partially) validating our underlying assumption.

\input{tables_and_figures/non_axis_align}
\input{tables_and_figures/repsize_vs_accuracy_pca}
\input{tables_and_figures/imagenet_pca_grad}

\subsection{Selecting Features from the Upper or Lower Tail of $A$}
We discussed in the Discriminative Feature Section that we select discriminative features by increasing the lower limits (from 0 percentile) and decreasing the upper limits (from 100 percentile). In Figure \ref{fig:upper_lower_tail}, we compare discriminative features with features selected from lower tail of $A$ (lower percentile fixed at 0) and the upper tail (upper percentile fixed at 100). Our discriminative features outperform both selection methods up to $60\%$ reduction in representation size. 

\input{tables_and_figures/upper_lower_tail}

\subsection{Ablation on $\lambda_1$ and $\lambda_2$}
In this section, we discuss how we can perform a hyper-parameter search on $\lambda_1$ and $\lambda_2$ to find the best performing pair of values. We take the baseline of SimCLR trained on ImageNet-1K and further train this model under the setup outlined in the Experimental Setup Section. We train keeping both $\lambda_1 = \lambda_2 = 0$ and run experiments by gradually increasing $\lambda_1$ to find the best performing value. Next, we search over $\lambda_2$ keeping the best performing value of $\lambda_1$. In these experiments we find that $\lambda_1 = \lambda_2 = 10^{-4}$ is the best performing pair. We find that this pair shows improved performance across most experiments. Due to lack of resources, we do not heavily tune these hyper-parameters in our experiments, however, we can expect improved performance if tuning is performed.

\input{tables_and_figures/search}

\subsection{Q-Score on Supervised Learning}
We note that we select discriminative features and compute Q-Score on self-supervised representations without using any label information. Thus, our study to show correlation between Q-score and classification outcome is non-trivial since self-supervised models learn without labels.
Nevertheless, we have included an experiment in Figure \ref{fig:supervised}, where we analyze Q-score as a predictor of classification outcome (correct vs incorrect) on supervised ResNet-18 (ImageNet-100) and ResNet-50 (ImageNet-1K) representations as well as their robust versions (l2 threat model). Self-supervised representations generally perform better than supervised representations on Q-score indicating that the representational properties we have identified may be mainly prominent in self-supervised learning. We observe that non-robust supervised ResNet shows lower AUROC and AUPRC compared to robust ResNet on both ImageNet-100 and ImageNet-1K setups. This is in line with observations in \cite{engstrom2020adversarial} and \cite{singla2021salient} that show that robust models provide better axis-alignment of features.

\input{tables_and_figures/supervised}



\input{tables_and_figures/avg_conf_Li}
\subsection{Q-Score and Classification Confidence} 
In Figure \ref{fig:avg_conf_Li}, we plot the mean of $|L_i|$ (left), i.e., number of highly activating features in the $i^{th}$ sample, and the mean linear classification confidence (right) over the population for each self-supervised model pre-trained with and without Q-Score regularization. We observe an increase in the average number of highly activating features ($L_i$) and as a result, an improvement in classification confidence, due to more enhanced features. 

\subsection{More Gradient Heatmaps of SimCLR}
In Figures \ref{fig:more_grad_cams_ski_mask}, \ref{fig:more_grad_cams_park_bench}, \ref{fig:more_grad_cams_head_cabbage} and \ref{fig:more_grad_cams_dutch_oven}, we plot more heatmaps of highly and lowly activating features of SimCLR for 4 different ImageNet-1K classes. We observe that the highly activating features correspond to unique physical properties that are correlated with the ground truth, whereas, lowly activating features, map to spurious portions that do not contribute to useful information.
\input{tables_and_figures/more_grad_cams}

%% file: tables_and_figures/representation_space.tex
\begin{figure}[h]
    \centering
    \subfigure{\includegraphics[width = 0.6\textwidth, trim = {0.4cm, 0.5cm, 5.8cm, 0.4cm}, clip]{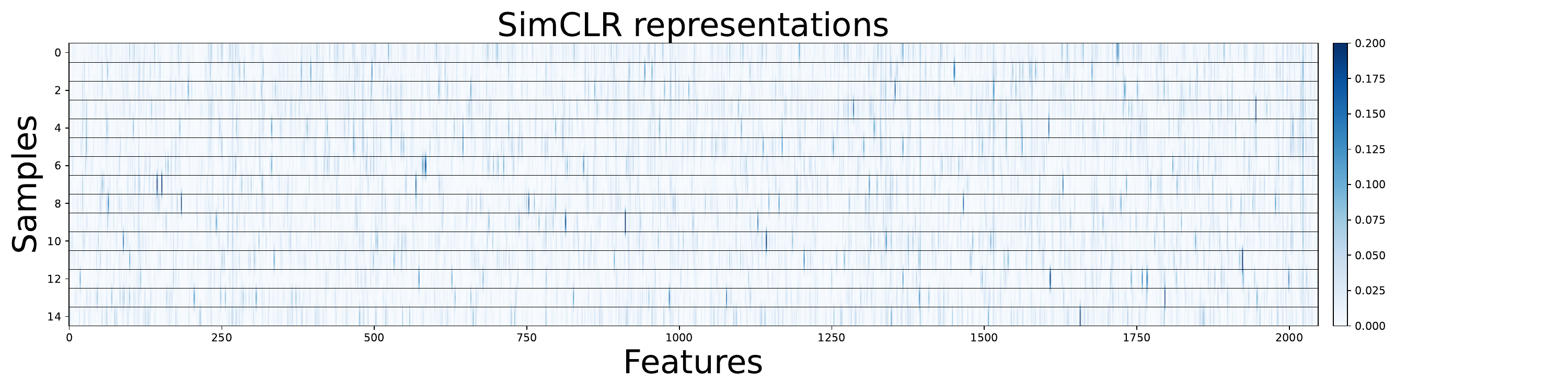}} \\
    \subfigure{\includegraphics[width = 0.2\textwidth, trim = {0.4cm, 0.5cm, 0.3cm, 0.4cm}, clip]{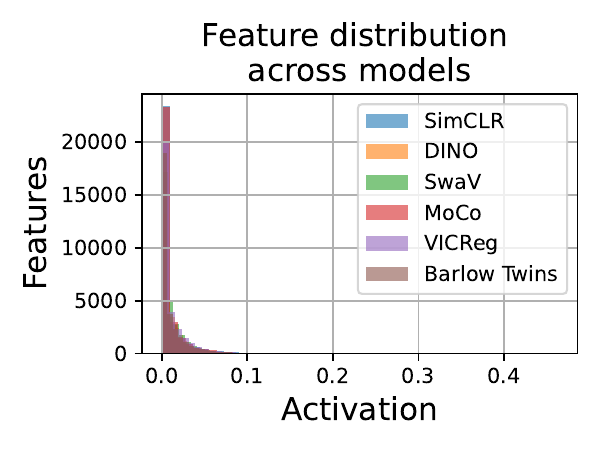}}
    \subfigure{\includegraphics[width = 0.2\textwidth, trim = {0.4cm, 0.5cm, 0.3cm, 0.4cm}, clip]{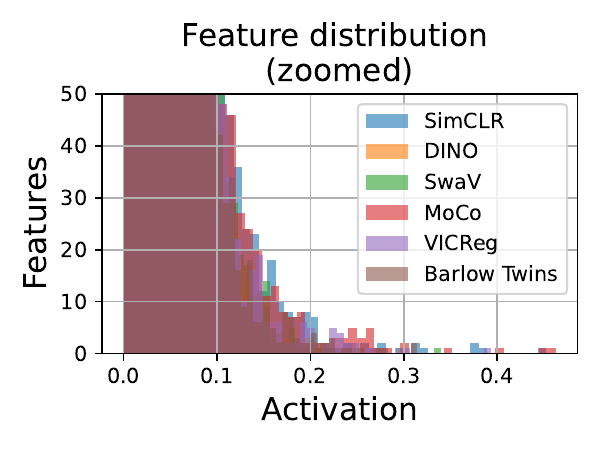}}
    \caption{\textbf{Visualizing the self-supervised representation space:} The top panel shows a heatmap of SimCLR representations of random ImageNet-1K samples. In the second panel, we plot the distribution of the features of the visualized samples for various models. We observe that representations are mostly sparse with a small number of strongly activated coordinates.}
    \label{fig:representation_space}
\end{figure}

%% file: tables_and_figures/results_non_imagenet.tex
\begin{table}[h]
    \centering
    \caption{\textbf{Linear classification performance with Q-Score regularization (more datasets):} Similar to Table \ref{tab:acc_baseline_vs_reg}, we tabulate our results on CIFAR-10, CIFAR-100 and STL-10}
    
    \resizebox{0.6\textwidth}{!}{
    
    \begin{tabular}{c|c|c|c|c|c|c|c|c|c}
    \toprule
    \multirow{2}{*}{\textbf{Model}} & \multicolumn{3}{c|}{\textbf{CIFAR-10}} & \multicolumn{3}{c|}{\textbf{CIFAR-100}} & \multicolumn{3}{c}{\textbf{STL-10}} \\
    & \textbf{Baseline} & \textbf{Lasso} & \textbf{Q-Score} & \textbf{Baseline} & \textbf{Lasso} & \textbf{Q-Score} & \textbf{Baseline} & \textbf{Lasso} & \textbf{Q-Score} \\
    \midrule
    \midrule
    
    SimCLR & 90.83 & 89.33 & \textbf{92.31} & 65.82 & 68.21 & \textbf{68.90} & 76.42 & 75.59 & \textbf{79.83} \\ 
    SwaV & 89.23 & 89.37 & \textbf{90.03} & 65.13 & 66.06 & \textbf{66.52} & 73.94 & 69.93 & \textbf{75.03} \\ 
    MoCo & 92.95 & 90.59 & \textbf{94.77} & 70.12 & 67.23 & \textbf{71.16} & 73.21 & 72.65 & \textbf{74.29} \\ 
    BYOL & 92.59 & 90.27 & \textbf{92.82} & 70.54 & 71.26 & \textbf{72.71} & 70.59 & 70.27 & \textbf{74.47} \\ 
    DINO & 89.54 & 89.57 & \textbf{89.85} & 66.82 & 65.52 & \textbf{67.49} & 68.36 & 69.29 & \textbf{69.38} \\ 
    SimSiam & 91.03 & 90.74 & \textbf{92.48} & 66.58 & 65.69 & \textbf{69.03} & 72.94 & 67.54 & \textbf{73.52} \\ 
    VICReg & 92.69 & 91.83 & \textbf{93.74} & 68.81 & 66.75 & \textbf{71.76} & 70.76 & 70.61 & \textbf{72.82} \\ 
    Barlow & 93.46 & 91.75 & \textbf{93.87} & 71.82 & 71.54 & \textbf{71.91} & 74.17 & 70.27 & \textbf{74.32} \\ 
    \bottomrule
    \end{tabular}
    }
    \label{tab:results_non_imagenet}
\end{table}

%% file: tables_and_figures/other_datasets.tex
\begin{table}[h]
\caption{\textbf{Transfer learning performance of various state-of-the-art self-supervised models trained on ImageNet-1K with and without Q-Score regularization:} We observe that fine-tuning with Q-Score regularization improves the average transfer accuracy on all self-supervised models.}
    \centering
    \resizebox{\textwidth}{!}{
    \begin{tabular}{c|c|c|c|c|c|c|c|c|c|c|c|c|c|c|c|c}
    \toprule
    \multirow{3}{*}{\textbf{Transfer}} & \multicolumn{2}{c|}{\textbf{SimCLR}} & \multicolumn{2}{c|}{\textbf{SwaV}} & \multicolumn{2}{c|}{\textbf{MoCo}} & \multicolumn{2}{c}{\textbf{BYOL}} & \multicolumn{2}{c|}{\textbf{DINO}} & \multicolumn{2}{c|}{\textbf{SimSiam}} & \multicolumn{2}{c|}{\textbf{VICReg}} & \multicolumn{2}{c}{\textbf{Barlow Twins}} \\
     \multirow{3}{*}{\textbf{Dataset}} & \multirow{2}{*}{\textbf{Baseline}} & \textbf{Q-Score} &  \multirow{2}{*}{\textbf{Baseline}} & \textbf{Q-Score} & \multirow{2}{*}{\textbf{Baseline}} & \textbf{Q-Score} &  \multirow{2}{*}{\textbf{Baseline}} & \textbf{Q-Score} & \multirow{2}{*}{\textbf{Baseline}} & \textbf{Q-Score} & \multirow{2}{*}{\textbf{Baseline}} & \textbf{Q-Score} & \multirow{2}{*}{\textbf{Baseline}} & \textbf{Q-Score} & \multirow{2}{*}{\textbf{Baseline}} & \textbf{Q-Score} \\
    & & \textbf{Regularized} & & \textbf{Regularized} & & \textbf{Regularized} & & \textbf{Regularized} & & \textbf{Regularized} & & \textbf{Regularized} & & \textbf{Regularized} & & \textbf{Regularized} \\
    \midrule
    \midrule
    CIFAR-10 & 70.13 & \textbf{70.55} & 71.27 & \textbf{72.42} & 72.39 & \textbf{73.26} & 71.36 & \textbf{72.99} & \textbf{72.62} & 70.33 & 72.62 & \textbf{74.29} & 73.32 & \textbf{73.98} & 71.23 & \textbf{73.83} \\
    CIFAR-100 & 40.23 & \textbf{40.70} & 42.52 & \textbf{42.69} & \textbf{45.70} & 44.11 & \textbf{45.92} & 45.36 & 43.32 & \textbf{46.45} & \textbf{45.85} & 42.93 & \textbf{42.20} & 42.04 & 41.01 & \textbf{45.18} \\
    STL-10 & 65.74 & \textbf{65.77} & 65.81 & \textbf{65.89} & 66.87 & \textbf{67.03} & 85.45 & \textbf{86.07} & \textbf{80.07} & 79.05 & 71.58 & \textbf{72.76} & 67.76 & \textbf{74.95} & \textbf{67.38} & 65.02 \\
    Aircraft & 62.38 & \textbf{68.13} & 63.86 & \textbf{73.08} & \textbf{69.34} & 67.59 & \textbf{63.76} & 62.14 & 63.9 & \textbf{73.9} & 66.5 & \textbf{72.75} & 68.22 & \textbf{65.69} & \textbf{64.3} & 63.12 \\
    Flowers & \textbf{88.12} & 85.19 & 86.35 & \textbf{86.99} & 89.19 & \textbf{89.61} & 87.37 & \textbf{89.59} & 87.97 & \textbf{89.81} &  \textbf{88.54} & 88.25 & 85.09 & \textbf{85.69} & 85.86 & \textbf{87.82} \\
    Food & 71.68 & \textbf{74.2} & \textbf{77.23} & 72.82 & 79.25 & \textbf{79.56} & 70.69 & \textbf{72.78} & 71.01 & \textbf{77.87} & \textbf{74.55} & 71.45 & \textbf{78.03} & 73.88 & 72.82 & \textbf{76.82} \\
    Cars & 51.61 & \textbf{54.26} & 50.74 & \textbf{53.05} & 54.37 & \textbf{54.87} & 51.83 & \textbf{54.85} & 51.22 & \textbf{51.91} & 50.92 & \textbf{50.64} & \textbf{52.29} & 50.44 & 51.5 & \textbf{53.51} \\
    DTD & 55.69 & \textbf{56.06} & 55.63 & \textbf{57.18} & 55.90 & \textbf{57.12} & 55.63 & \textbf{56.06} & 50.9 & \textbf{53.77} & 52.07 & \textbf{53.24} & 51.27 & \textbf{52.63} & \textbf{54.07} & 51.08 \\
    \midrule
    Average & 63.12 & \textbf{64.36} & 64.18 & \textbf{65.52} & 66.62 & \textbf{66.64} & 66.50 & \textbf{67.48} & 65.13 & \textbf{67.89} & 65.33 & \textbf{65.79} & 64.77 & \textbf{64.91}	& 63.52 & \textbf{64.55} \\
    
    \bottomrule
    \end{tabular}
    }

    \label{tab:acc_other_datasets}
\end{table}

%% file: tables_and_figures/grad_transfer.tex
\begin{figure}[h]
    \centering
    \subfigure{\includegraphics[width=0.23\textwidth]{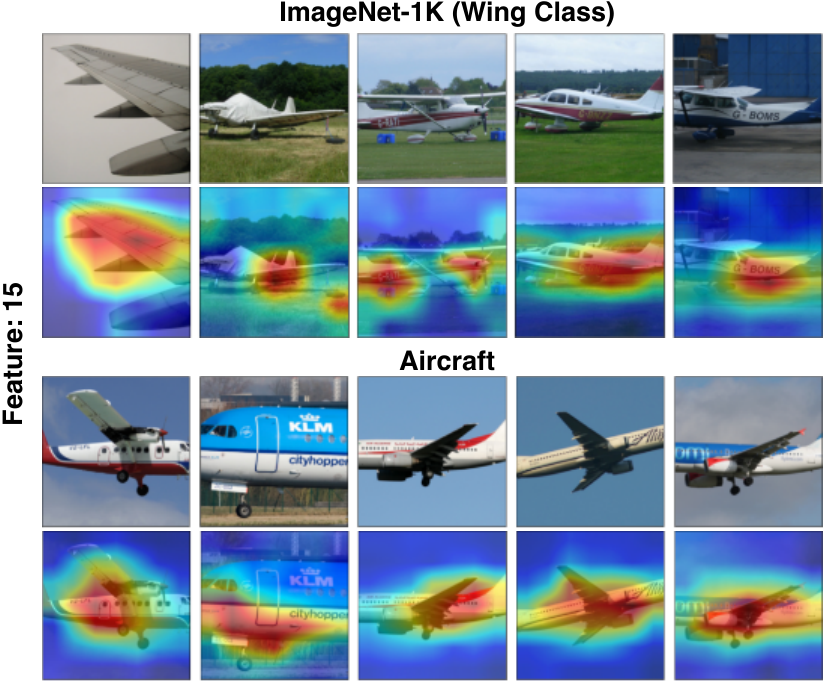}}
    \subfigure{\includegraphics[width=0.23\textwidth]{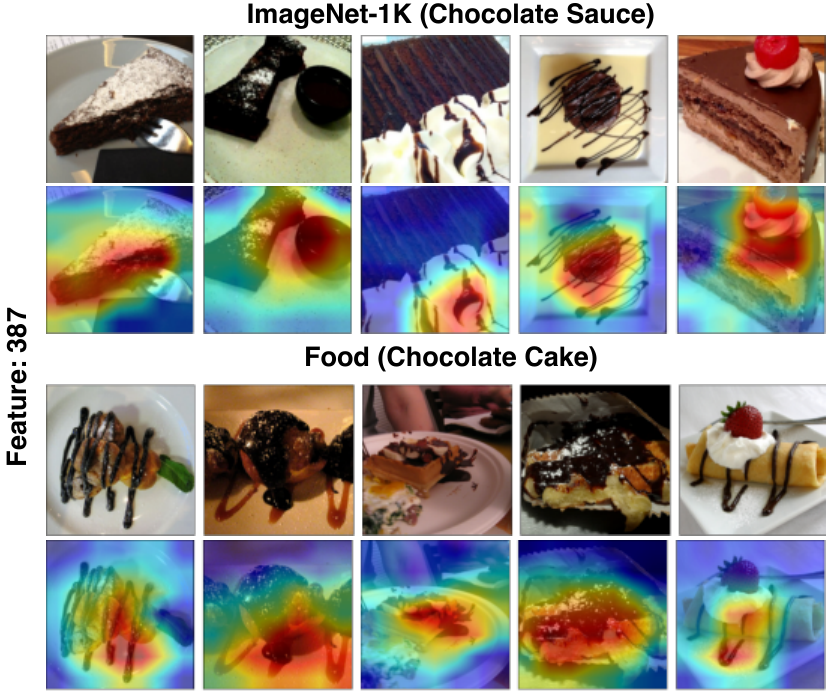}}
    \subfigure{\includegraphics[width=0.23\textwidth]{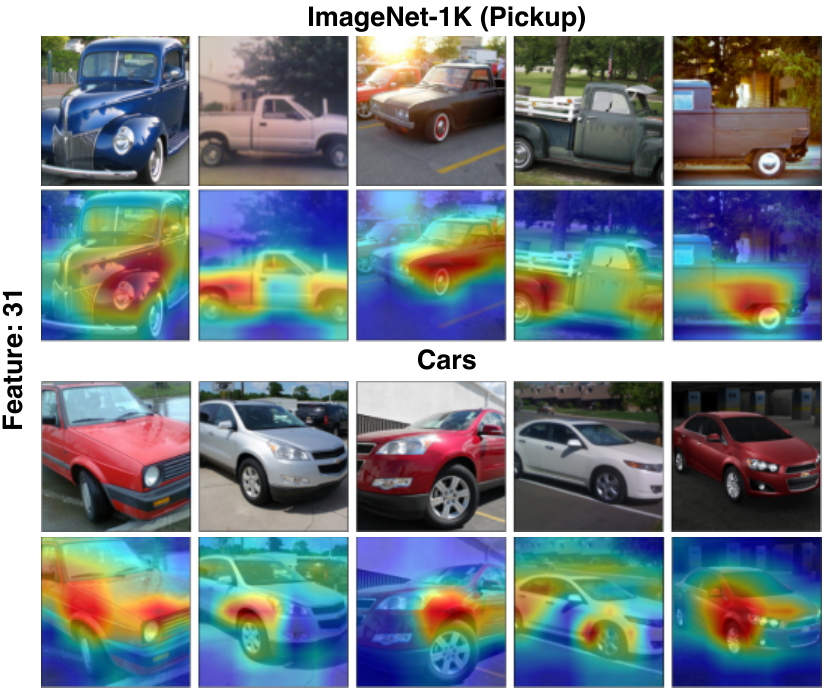}}
    \caption{\textbf{Discriminative features on unseen datasets:} We visualize the discriminative features discovered on ImageNet-1K classes on unseen datasets like Aircraft, Food and Cars. We observe that discriminative features correspond to the same physical attributes as the training data and are core and informative.}
    \label{fig:grad_transfer}
\end{figure}

%% file: tables_and_figures/correct_vs_incorrect_flowers.tex
\begin{figure*}[h]
    \centering
    \includegraphics[width=0.38\textwidth, trim={0cm 0cm 0cm
0cm}, clip]{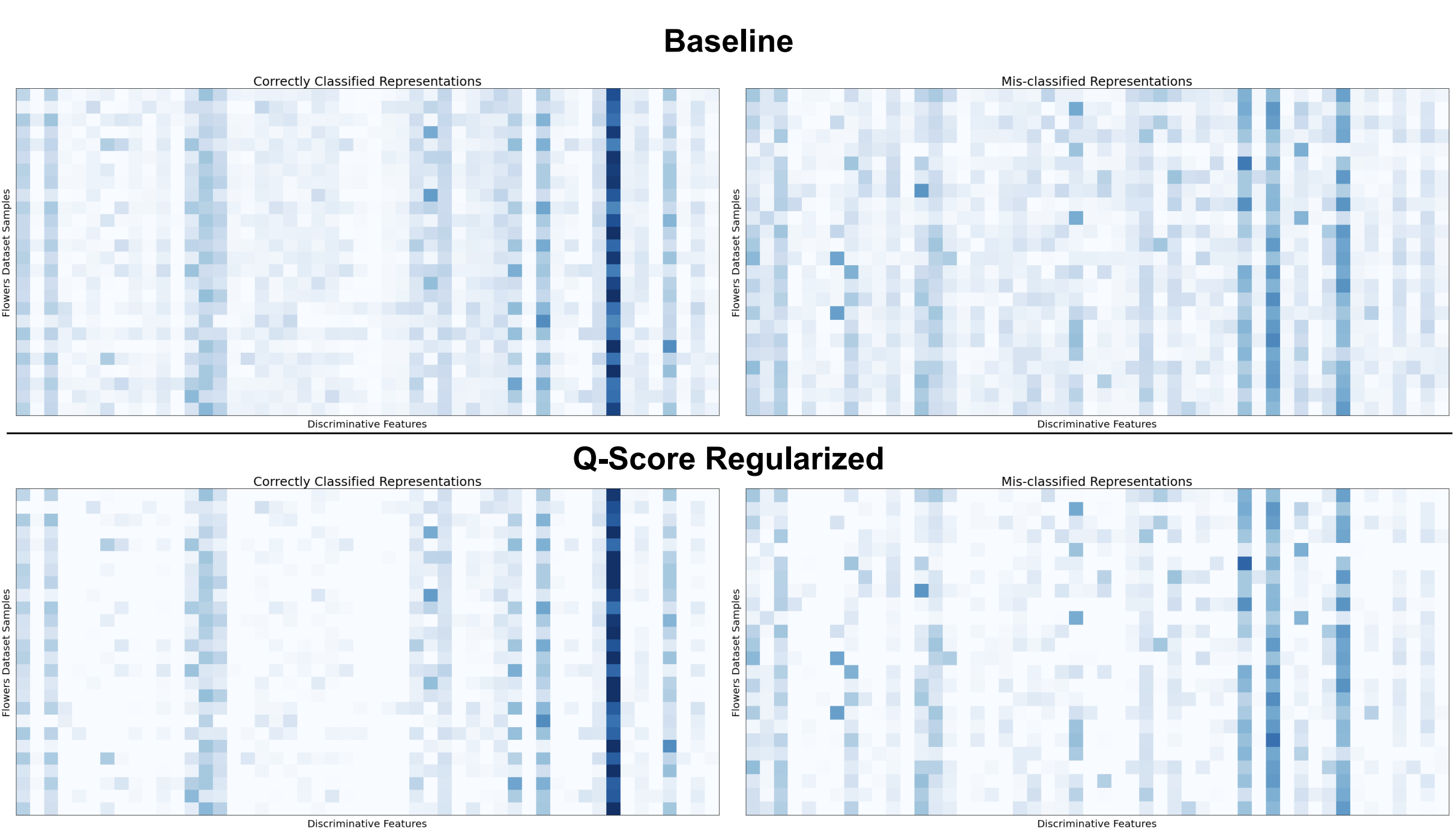}
    \caption{\textbf{Comparing correct and mis-classified representations in Flowers dataset:} In these heatmaps, we visualize the discriminative features of several Flowers dataset samples. In the top panel, we display the correct (left) and incorrect (right) classifications of SimCLR (trained on ImageNet-1K) and in the bottom panel, we visualize the same when pre-trained using Q-Score regularization. Similar to the observations in Figure \ref{fig:correct_vs_incorrect_latent}, we observe that the regularization enhances discriminative features, thereby leading to an improvement in performance. }
    \label{fig:correct_vs_incorrect_flowers}
\end{figure*}

%% file: tables_and_figures/non_axis_align.tex
\begin{figure}[h]
    \centering
    \includegraphics[width=0.15\textwidth]{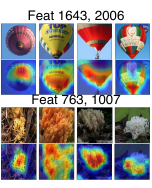}
    \caption{\textbf{Feature groups:} Groups of features (non-axis-aligned) can still correspond to meaningful physical concepts that are associated with class labels, therefore, can also be discriminative. }
    \label{fig:non_axis_align}
\end{figure}

%% file: tables_and_figures/repsize_vs_accuracy_pca.tex
\begin{figure*}[h]
    \centering
    \subfigure{\includegraphics[width=0.6\textwidth, trim = {0.4cm, 0.5cm, 0.3cm, 0.4cm}, clip]{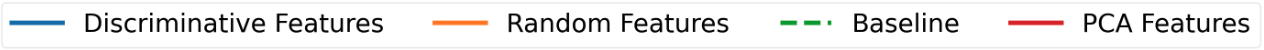}}\\
    \subfigure{\includegraphics[width=0.12\textwidth, trim = {0.4cm, 0.5cm, 0.4cm, 0.5cm}, clip]{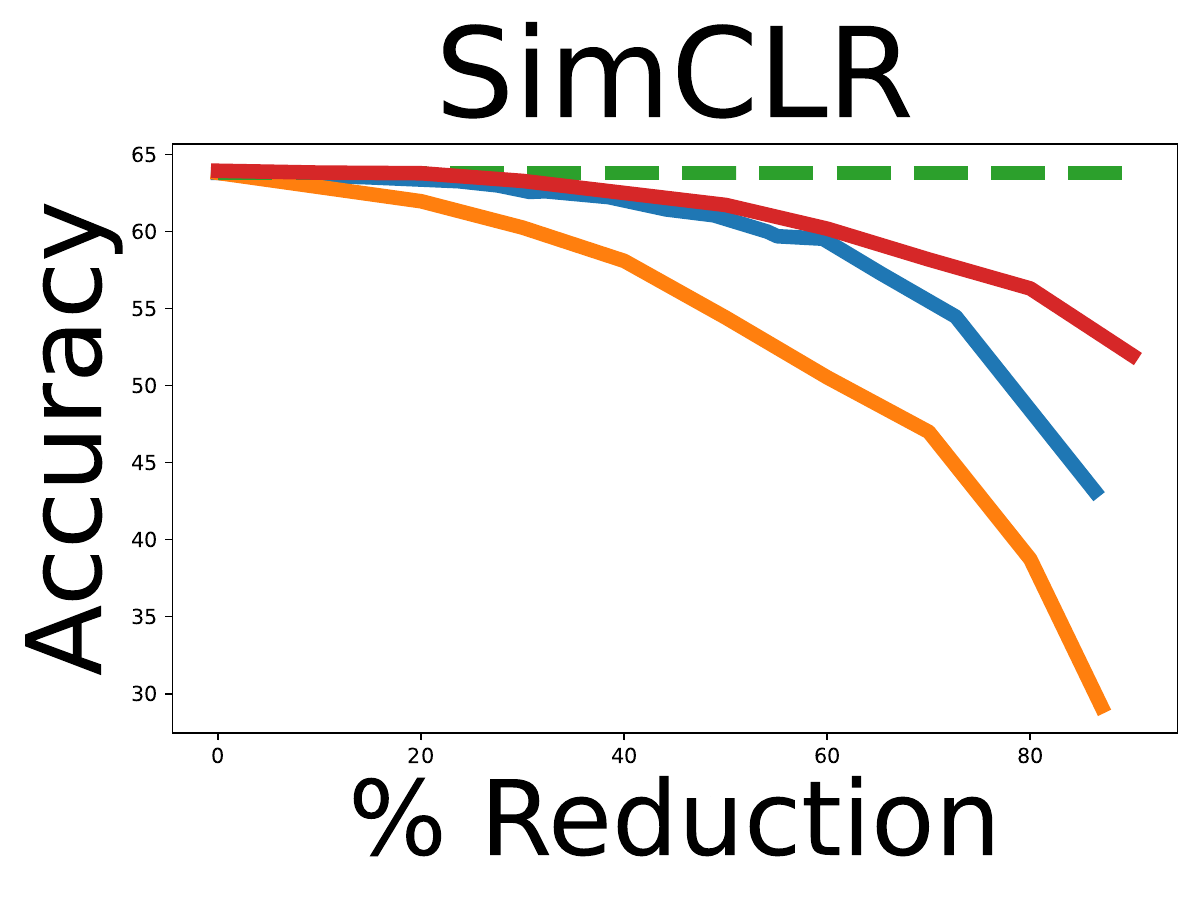}}
    \subfigure{\includegraphics[width=0.12\textwidth, trim = {0.4cm, 0.5cm, 0.4cm, 0.5cm}, clip]{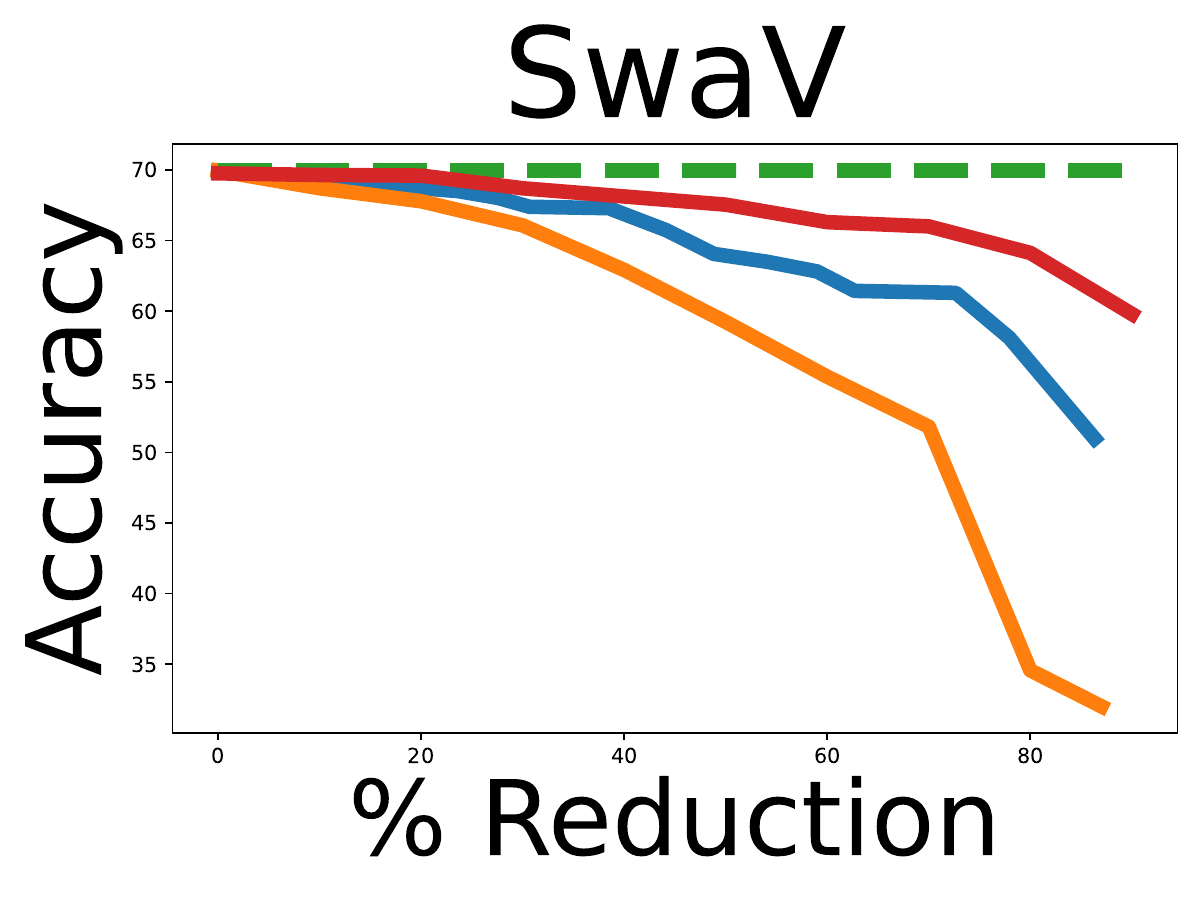}}
    \subfigure{\includegraphics[width=0.12\textwidth, trim = {0.4cm, 0.5cm, 0.4cm, 0.5cm}, clip]{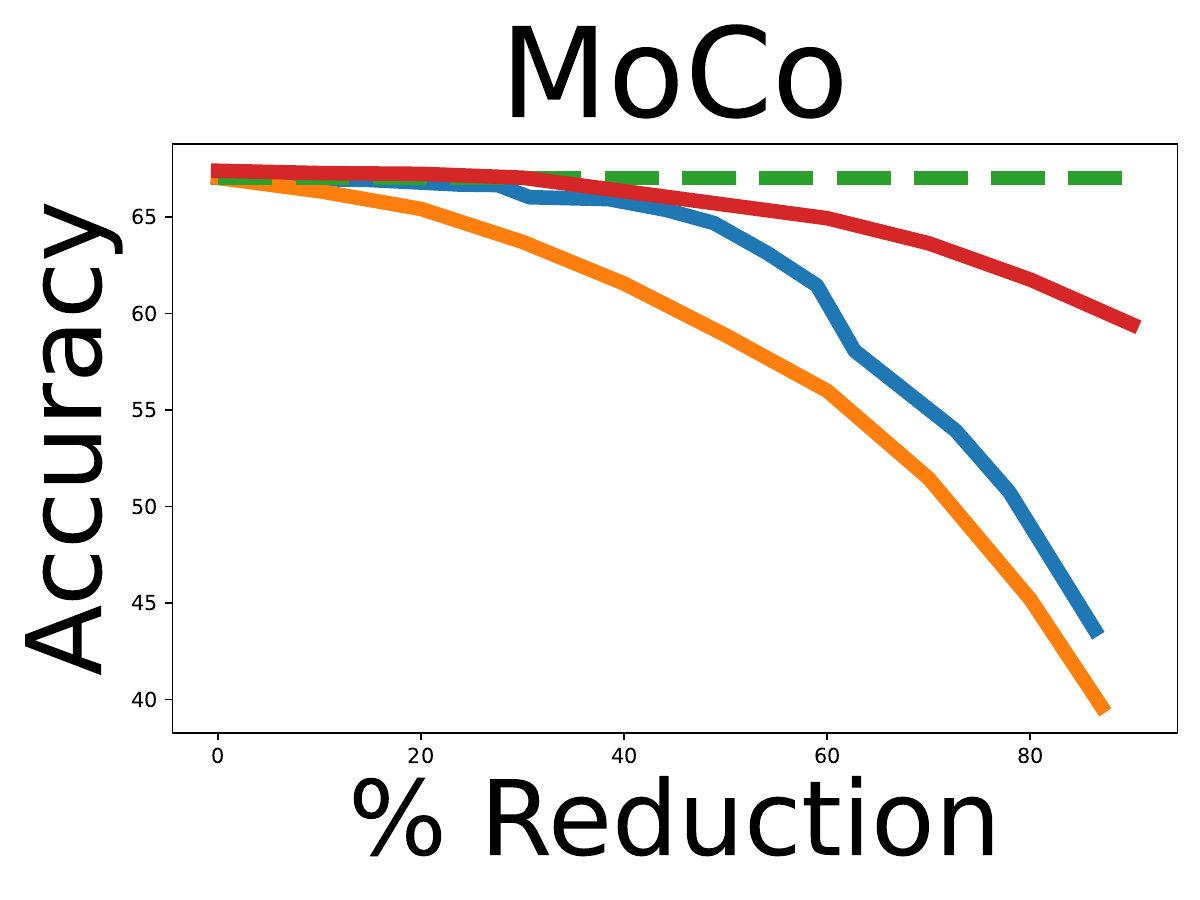}}
    \subfigure{\includegraphics[width=0.12\textwidth, trim = {0.4cm, 0.5cm, 0.4cm, 0.5cm}, clip]{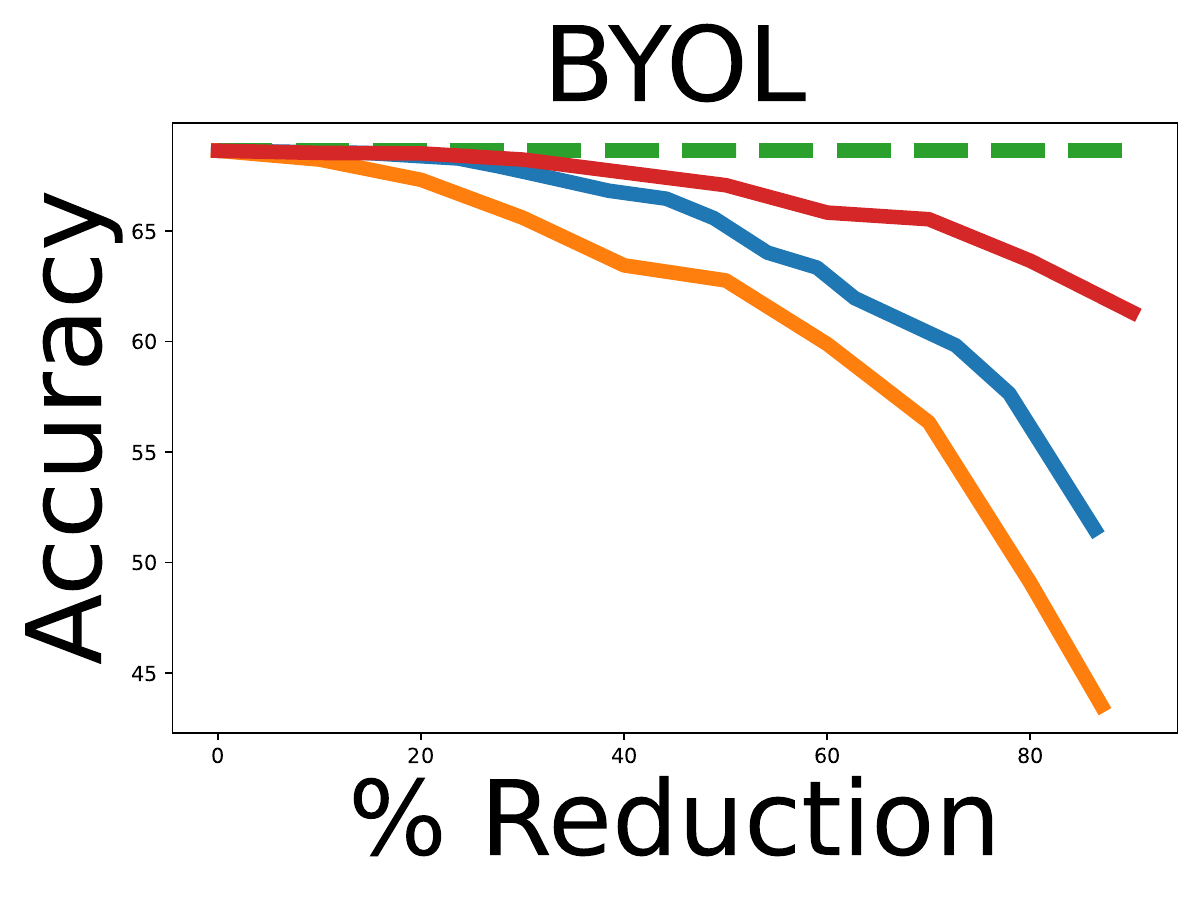}}
    \subfigure{\includegraphics[width=0.12\textwidth, trim = {0.4cm, 0.5cm, 0.4cm, 0.5cm}, clip]{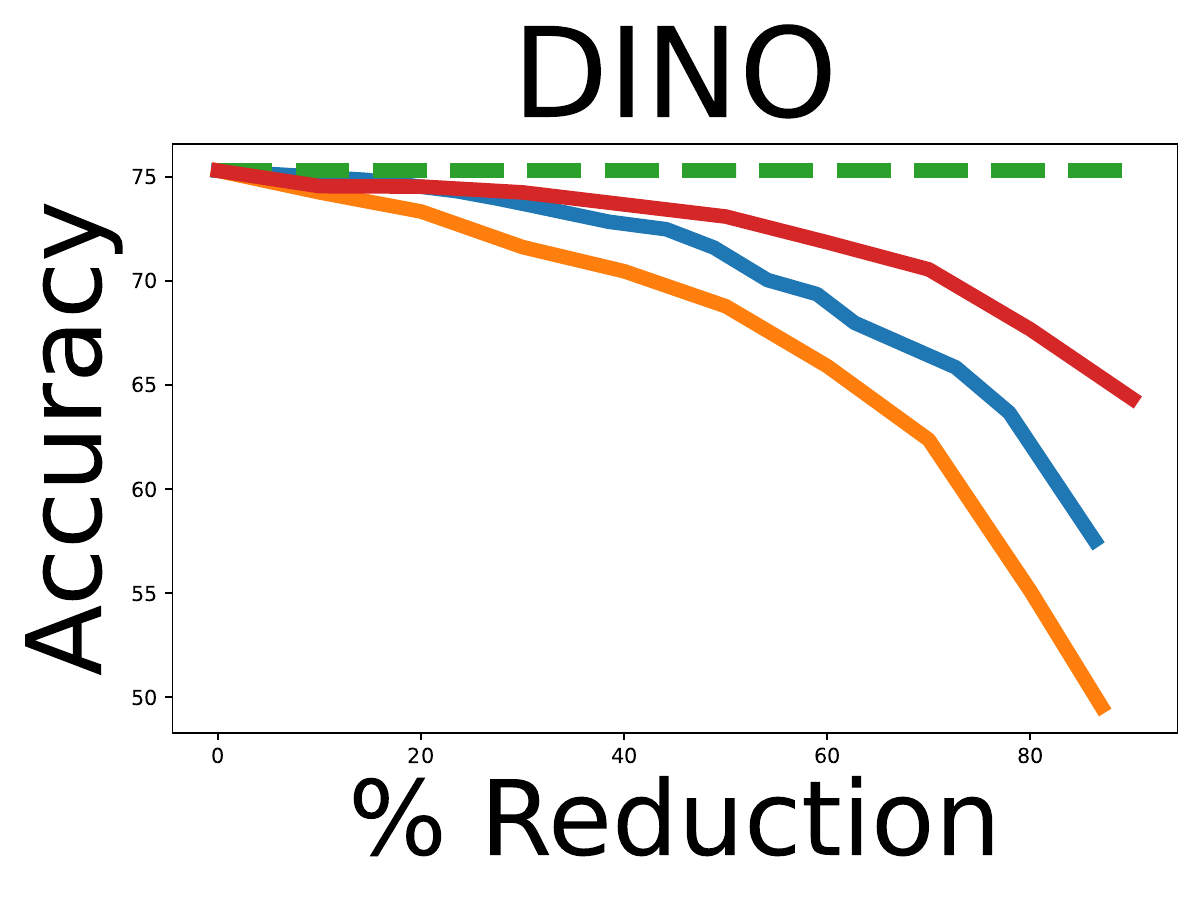}}
    \subfigure{\includegraphics[width=0.12\textwidth, trim = {0.4cm, 0.5cm, 0.4cm, 0.5cm}, clip]{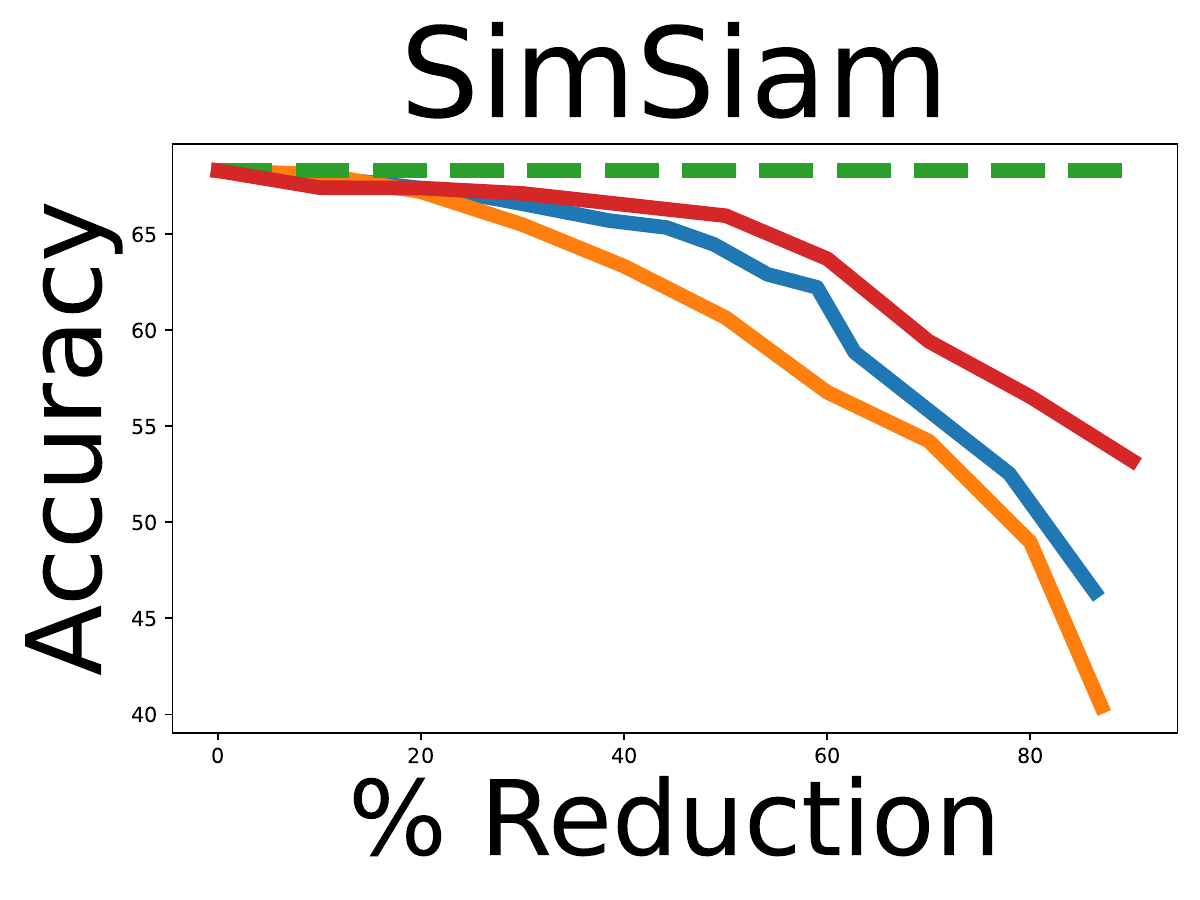}}
    \subfigure{\includegraphics[width=0.12\textwidth, trim = {0.4cm, 0.5cm, 0.4cm, 0.5cm}, clip]{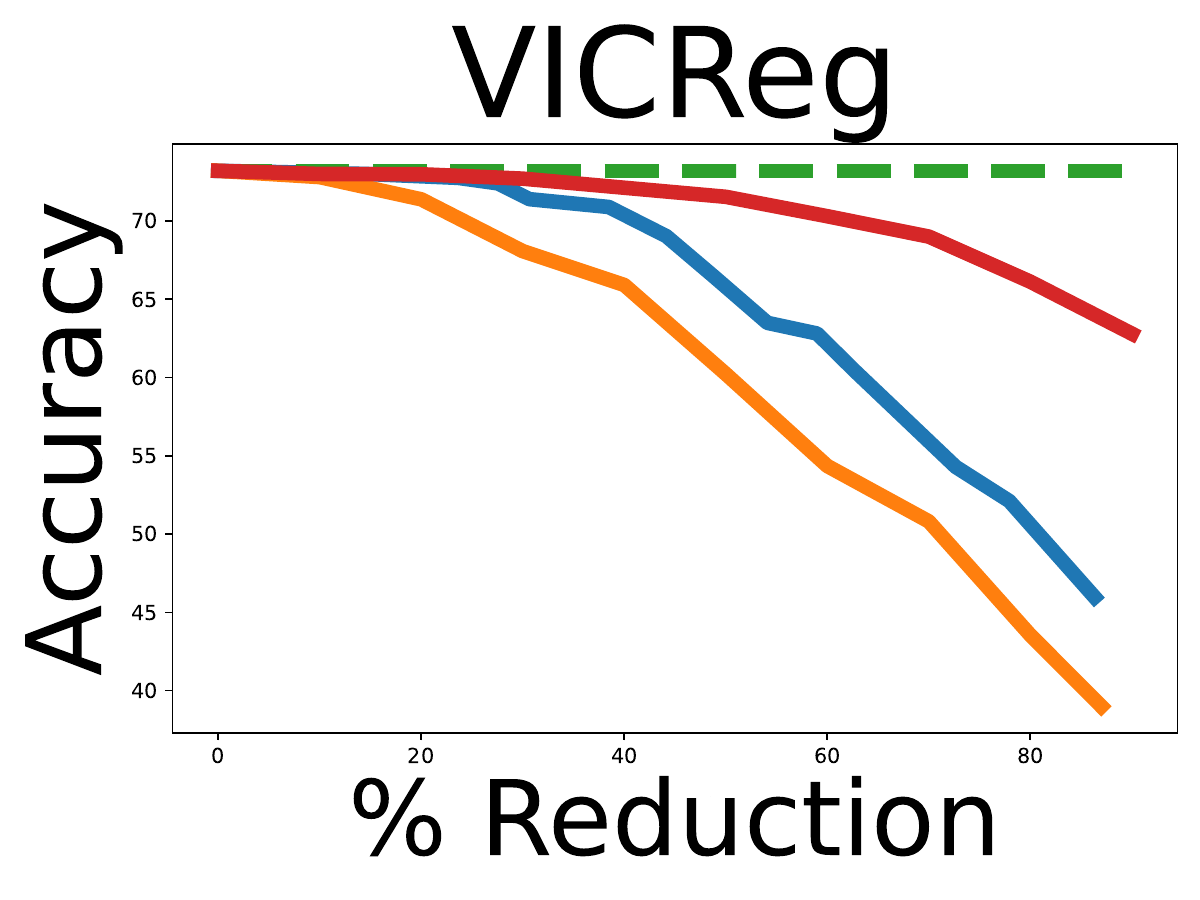}}
    \subfigure{\includegraphics[width=0.12\textwidth, trim = {0.4cm, 0.5cm, 0.4cm, 0.5cm}, clip]{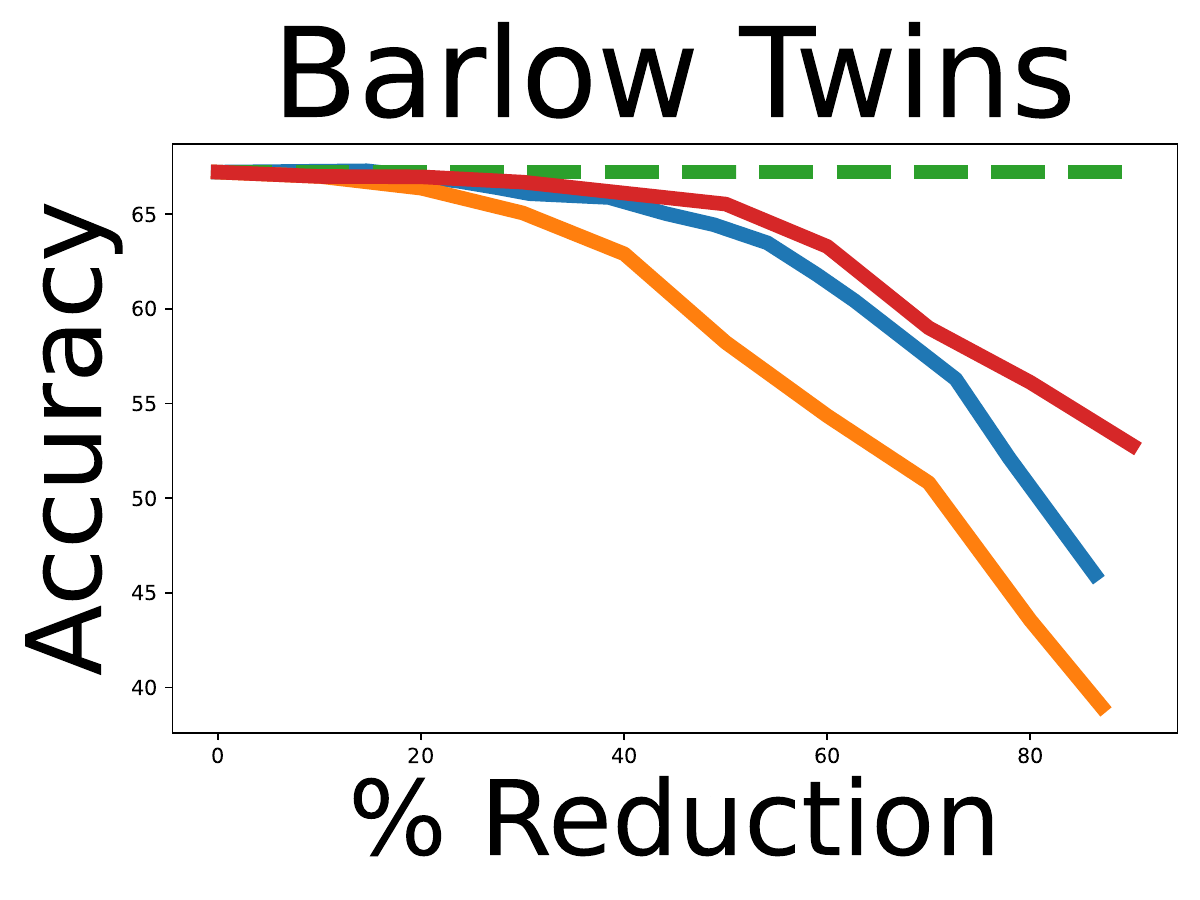}}
    \caption{\textbf{Linear classification accuracy on discriminative features:} Similar to Figure \ref{fig:repsize_vs_accuracy}, we compare discriminative and random features to PCA features of matching sizes. Discriminative features also match the performance of PCA features to a certain extent showing that features can be considered as axis-aligned.}
    \label{fig:repsize_vs_accuracy_pca}
\end{figure*}

%% file: tables_and_figures/imagenet_pca_grad.tex
\begin{figure}[h]
    \centering
    \includegraphics[width = 0.5\textwidth]{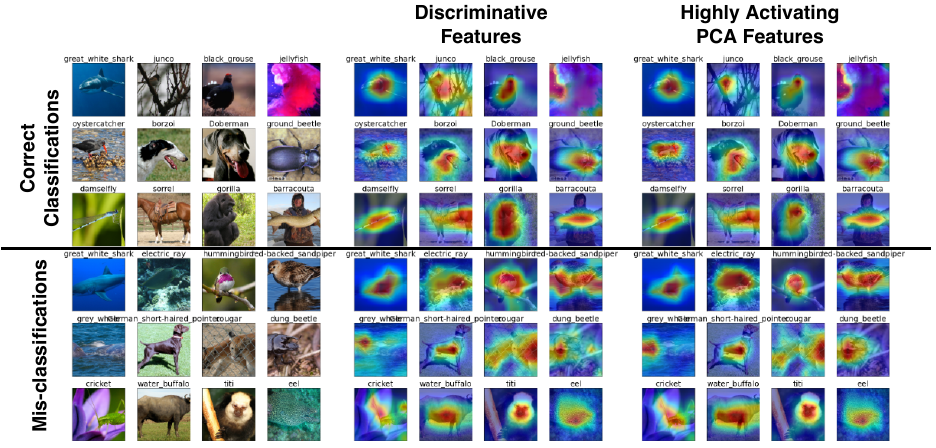}
    \caption{\textbf{Comparing gradient heatmaps of discriminative features and PCA features:} In this figure, we plot the gradient heatmaps of the discriminative features of correct and incorrect classifications on ImageNet-1K trained on SimCLR. We also plot the discriminative PCA features for the same images. We observe that both sets of features activate the same portions of the images meaning that discriminative features can we viewed as axis-aligned.}
    \label{fig:imagenet_pca_grad}
\end{figure}

%% file: tables_and_figures/upper_lower_tail.tex
\begin{figure}[h]
    \centering
    \includegraphics[width=0.2\textwidth]{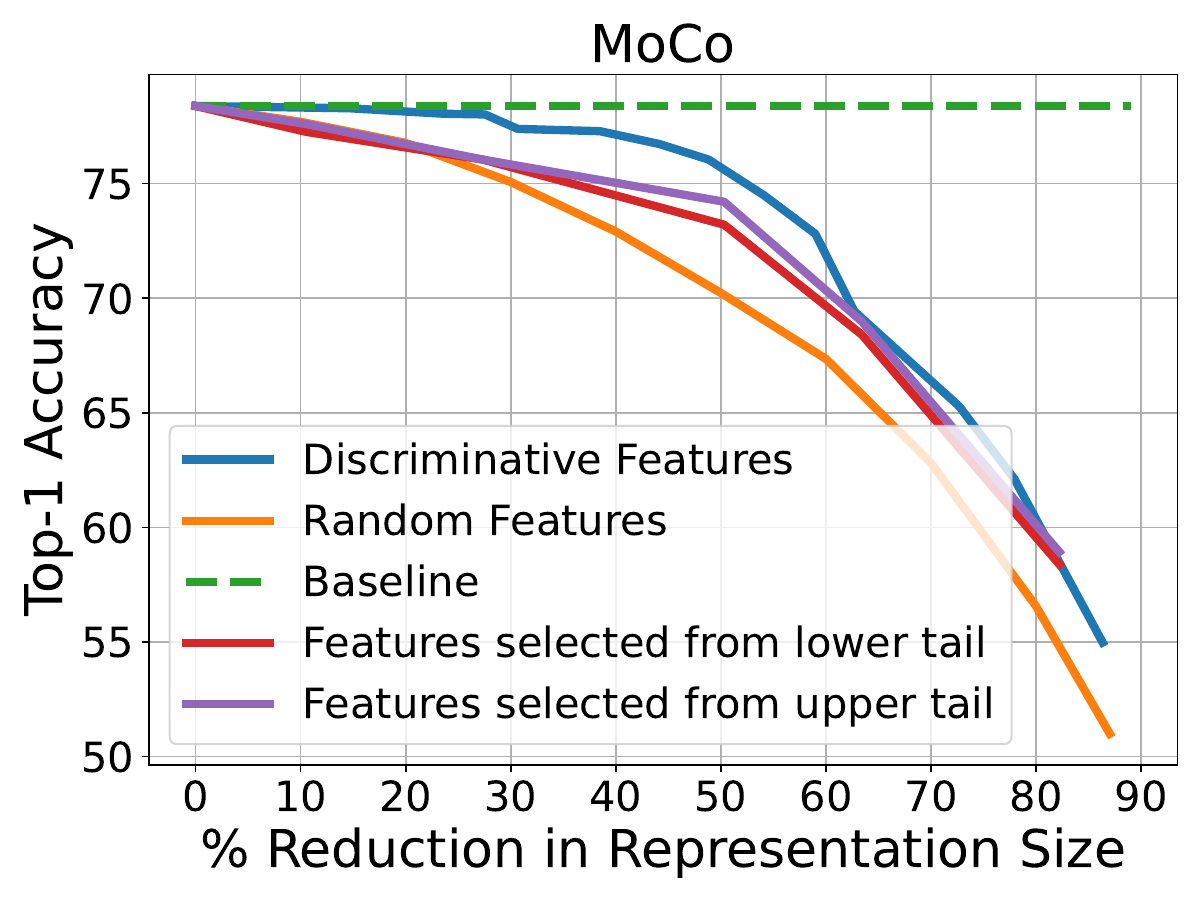}
    \caption{\textbf{Selecting discriminative features:} We additionally include the results on MoCo, ImageNet-100 when selecting features either from the lower or upper tail of $A$ which perform worse than the middle portion of $A$. }
    \label{fig:upper_lower_tail}
\end{figure}

%% file: tables_and_figures/search.tex
\begin{figure}[h]
    \centering
    \subfigure{\includegraphics[width=0.2\textwidth]{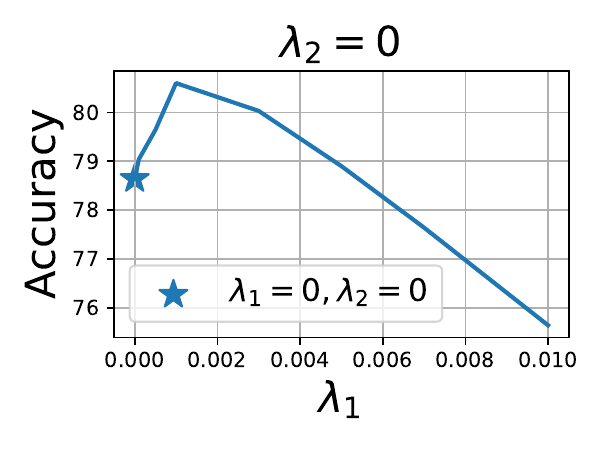}}
    \subfigure{\includegraphics[width=0.2\textwidth]{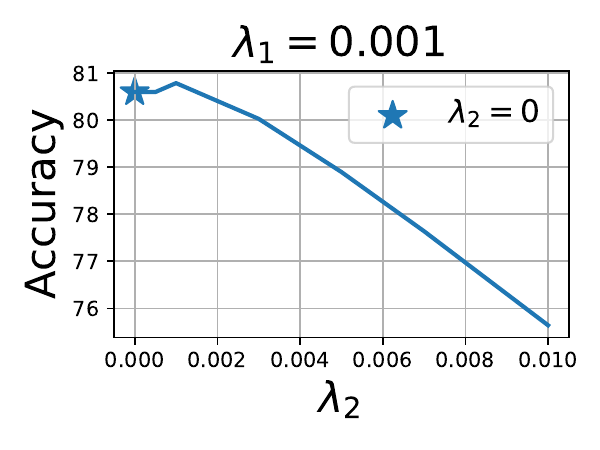}}
    \caption{\textbf{Hyper-parameter Search on $\lambda_1$ and $\lambda_2$:} We set $\lambda_2 = 0$ and search across various values of $\lambda_1$ to find the best performing experiment. Next, we set $\lambda_1$ to the best performing value and search over $\lambda_2$.}
    \label{fig:search}
\end{figure}

%% file: tables_and_figures/supervised.tex
\begin{figure}
    \centering
    \subfigure{\includegraphics[width=0.3\textwidth]{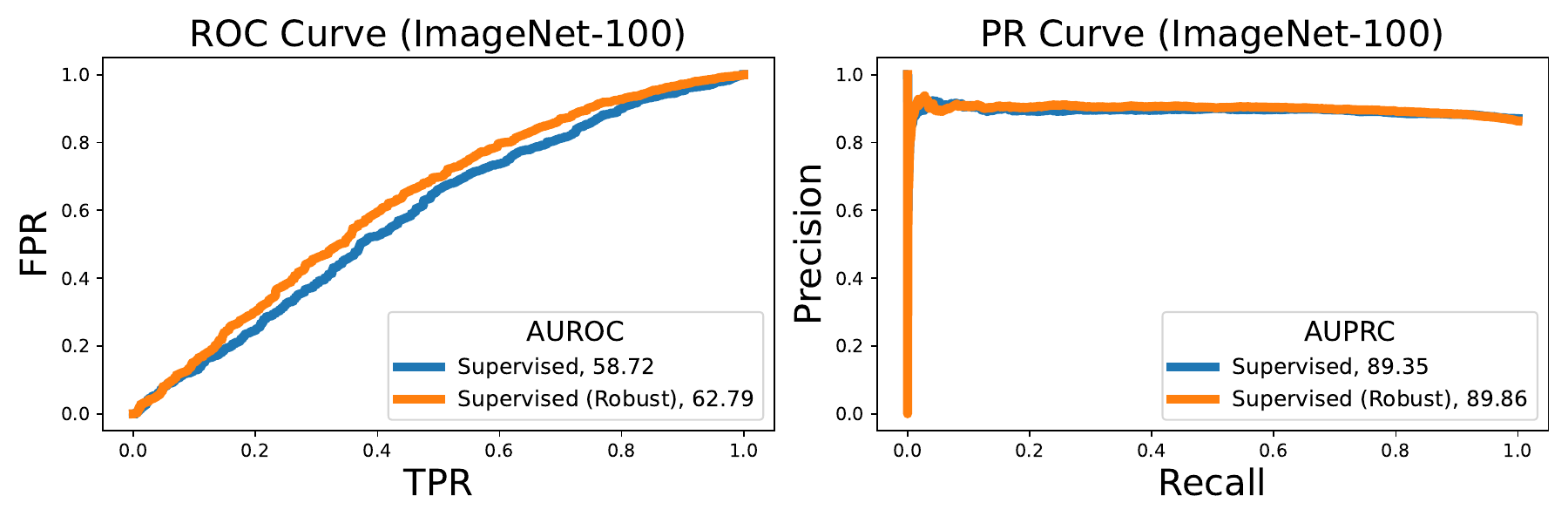}}
    \subfigure{\includegraphics[width=0.3\textwidth]{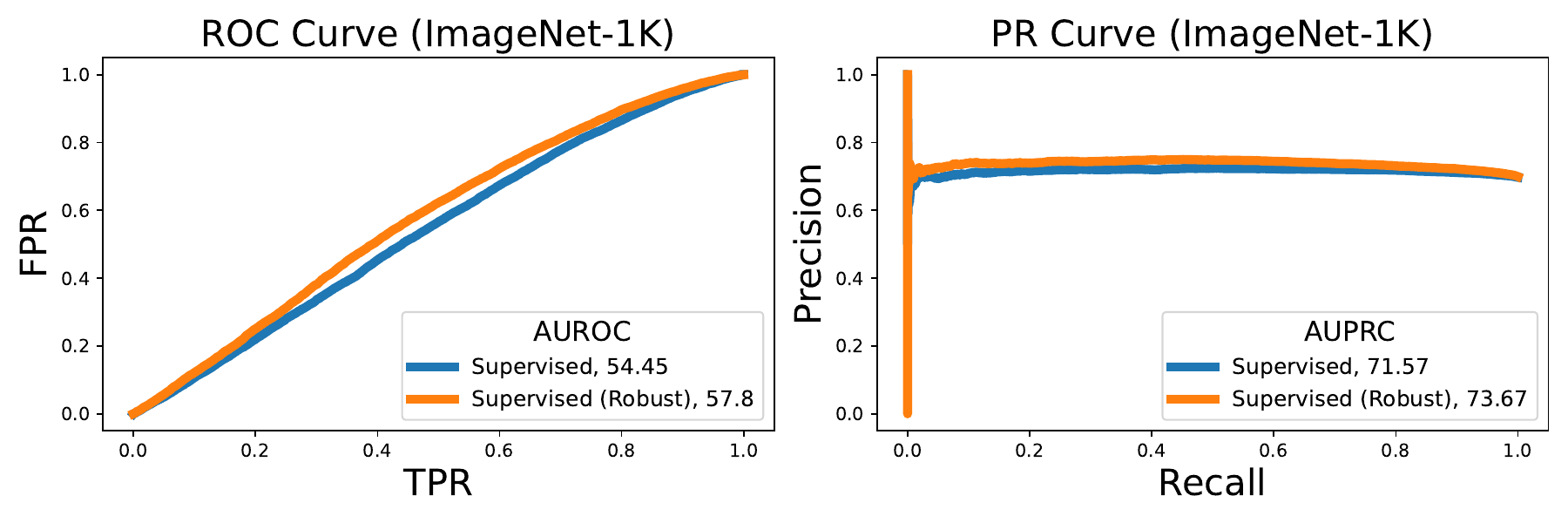}}
    \caption{\textbf{Precision-Recall and ROC curves of Q-Score on supervised setups:} In the first two plots, we compute the ROC and PR curves (similar to Figure \ref{fig:auc_plots}) of Q-score on the representations of a supervised ResNet-18 model and a robust ResNet-18 trained on ImageNet-100. In the last two plots, we show the same for ResNet-50 trained on ImageNet-1K. We observe that robust ResNet performs better for Q-score when used as a predictor for correct or mis-classified representations.}
    \label{fig:supervised}
\end{figure}

%% file: tables_and_figures/avg_conf_Li.tex
\begin{figure}
    \centering
    \subfigure{\includegraphics[width=0.23\textwidth]{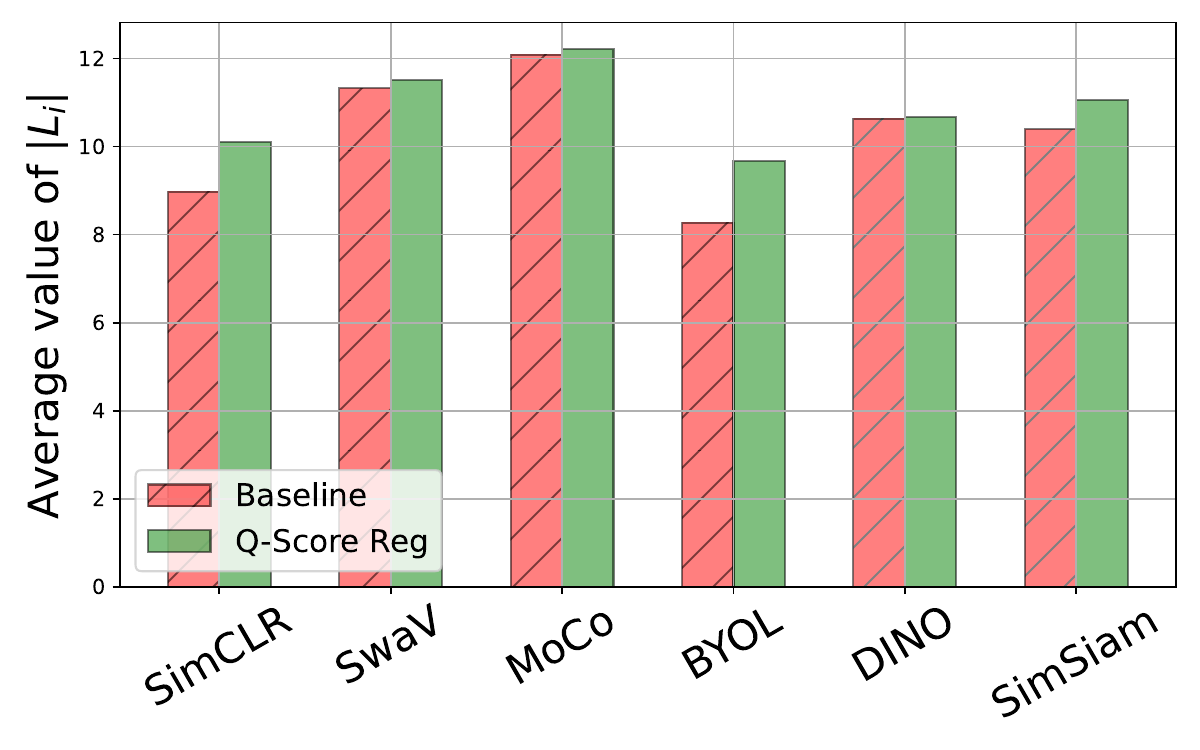}}
    \subfigure{\includegraphics[width=0.23\textwidth]{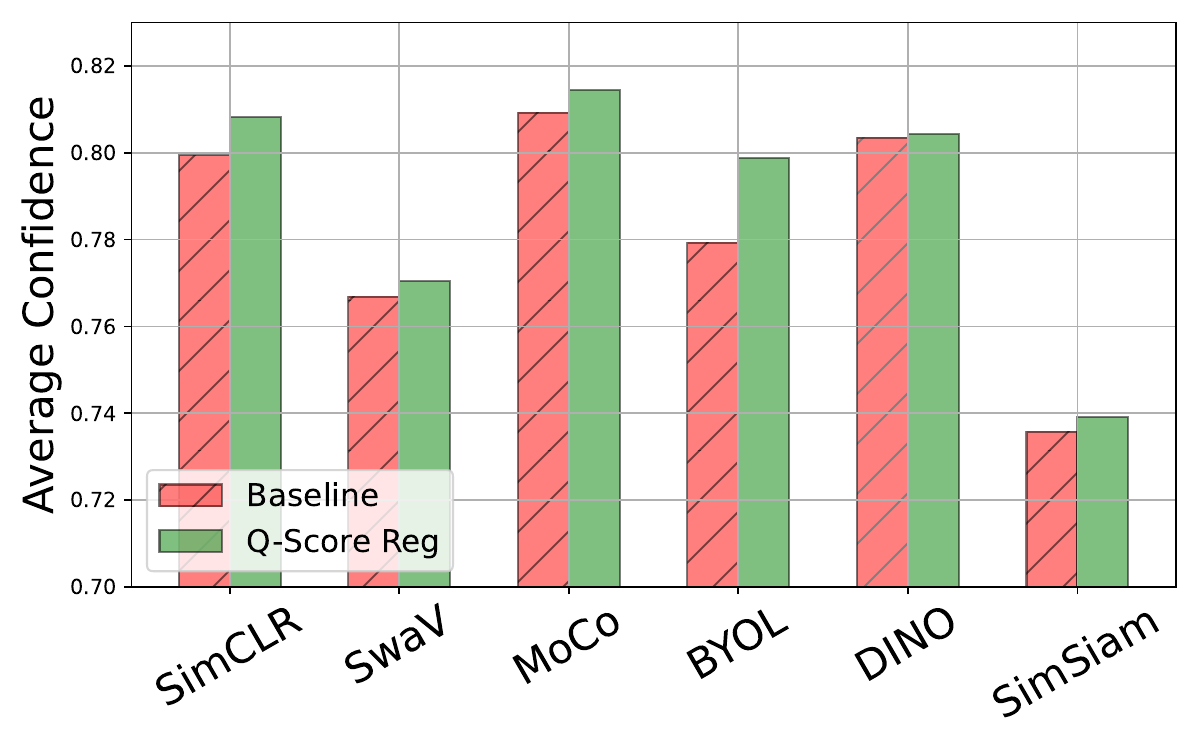}}
    \caption{\textbf{Average $|L_i|$ (left) and classification confidence (right) before and after regularization:} On the left we plot the average value of $|L_i|$ (number of highly activating features) and on the right we plot the average classification confidence over the population of ImageNet-1K. We observe that both the number of highly activating features and classification confidence consistently improve on every self-supervised baseline with Q-Score regularization. This improvement is due to the nature of Q-Score regularization which maximizes highly activating discriminative features over the course of pre-training leading to a higher number of such features and improved classification confidence. }
    \label{fig:avg_conf_Li}
\end{figure}

%% file: tables_and_figures/more_grad_cams.tex
\begin{figure*}
    \centering
    
    \subfigure{\includegraphics[width = 0.8\textwidth, trim={0cm 0cm 0cm
0cm}, clip]{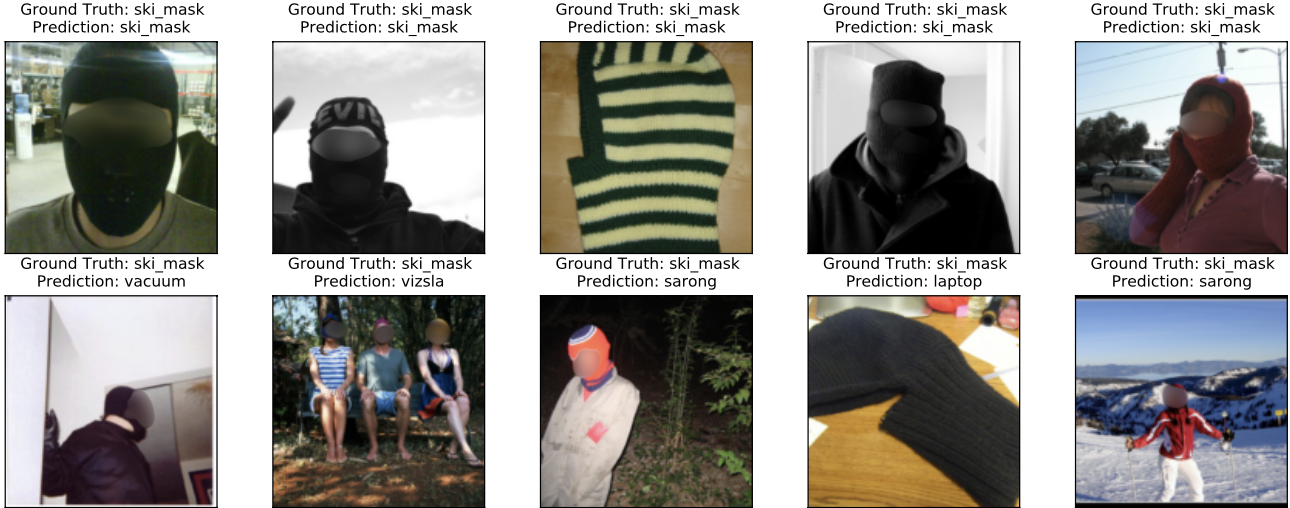}}
    
    \subfigure{\includegraphics[width = 0.8\textwidth, trim={0cm 0cm 0cm
0cm}, clip]{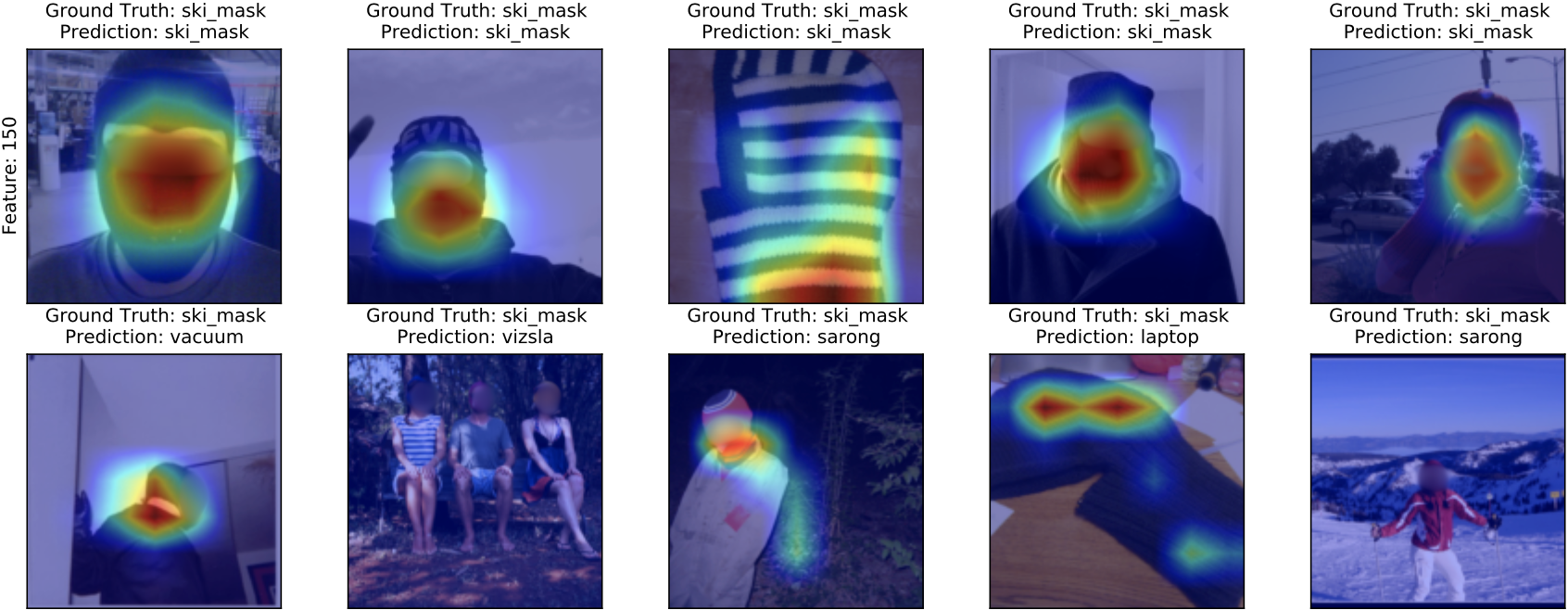}}

    \subfigure{\includegraphics[width = 0.8\textwidth, trim={0cm 0cm 0cm
0cm}, clip]{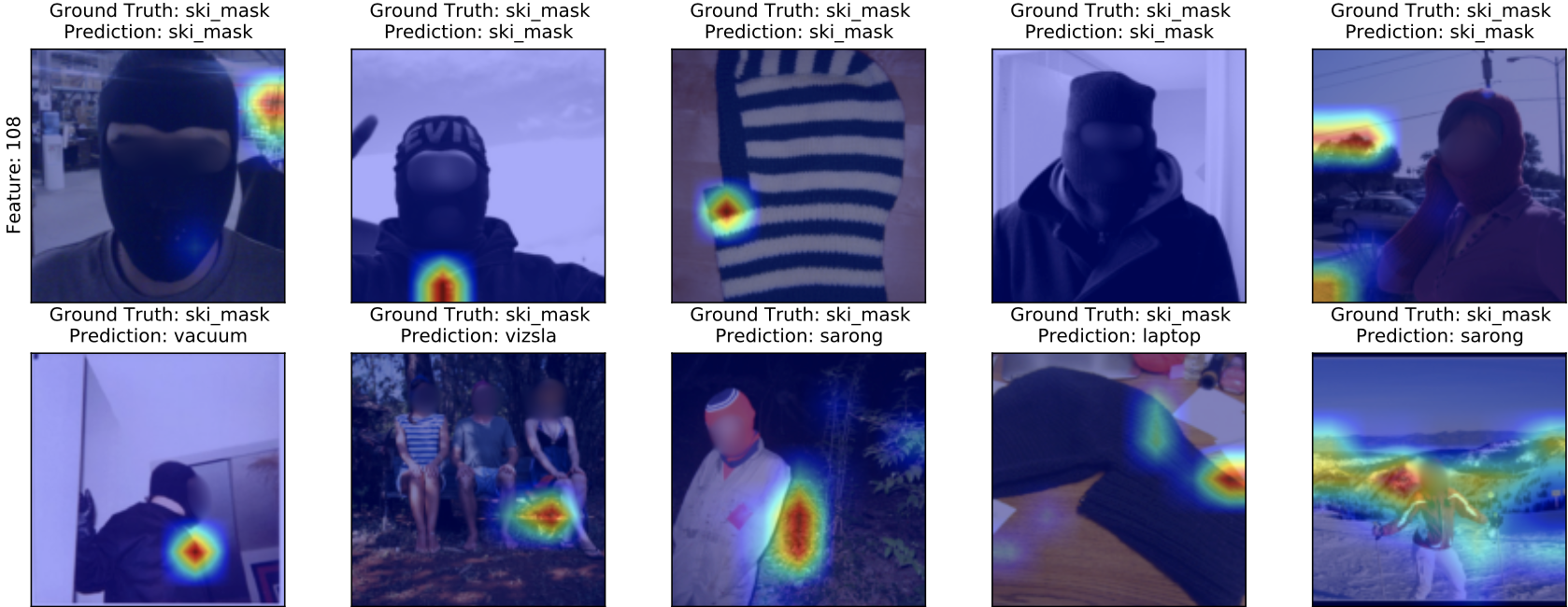}}
    
    \caption{\textbf{Heatmaps of discriminative and lowly activating features of SimCLR (Class - Ski Mask)}: We plot the gradient heat maps of the top activating discriminative feature (by magnitude) for the given class and a lowly activating feature of the same class. We observe that discriminative features are more correlated with ground truth labels compared to lowly activating features in both correct and incorrect classification. The discriminative feature in correct classifications correspond to a unique physical attribute that may not exist (or be obfuscated) in mis-classified images.}
    \label{fig:more_grad_cams_ski_mask}
\end{figure*}

\begin{figure*}
    \centering
    \subfigure{\includegraphics[width = 0.8\textwidth, trim={0cm 0cm 0cm
0cm}, clip]{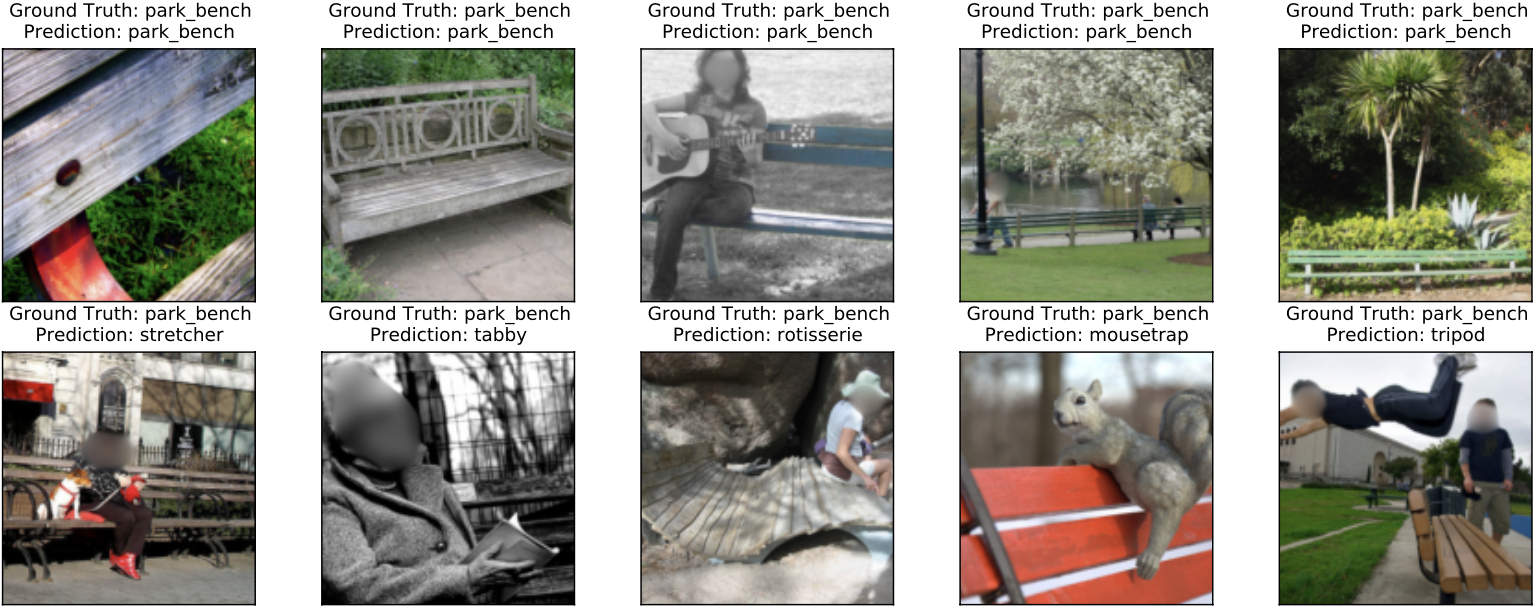}}

    \subfigure{\includegraphics[width = 0.8\textwidth, trim={0cm 0cm 0cm
0cm}, clip]{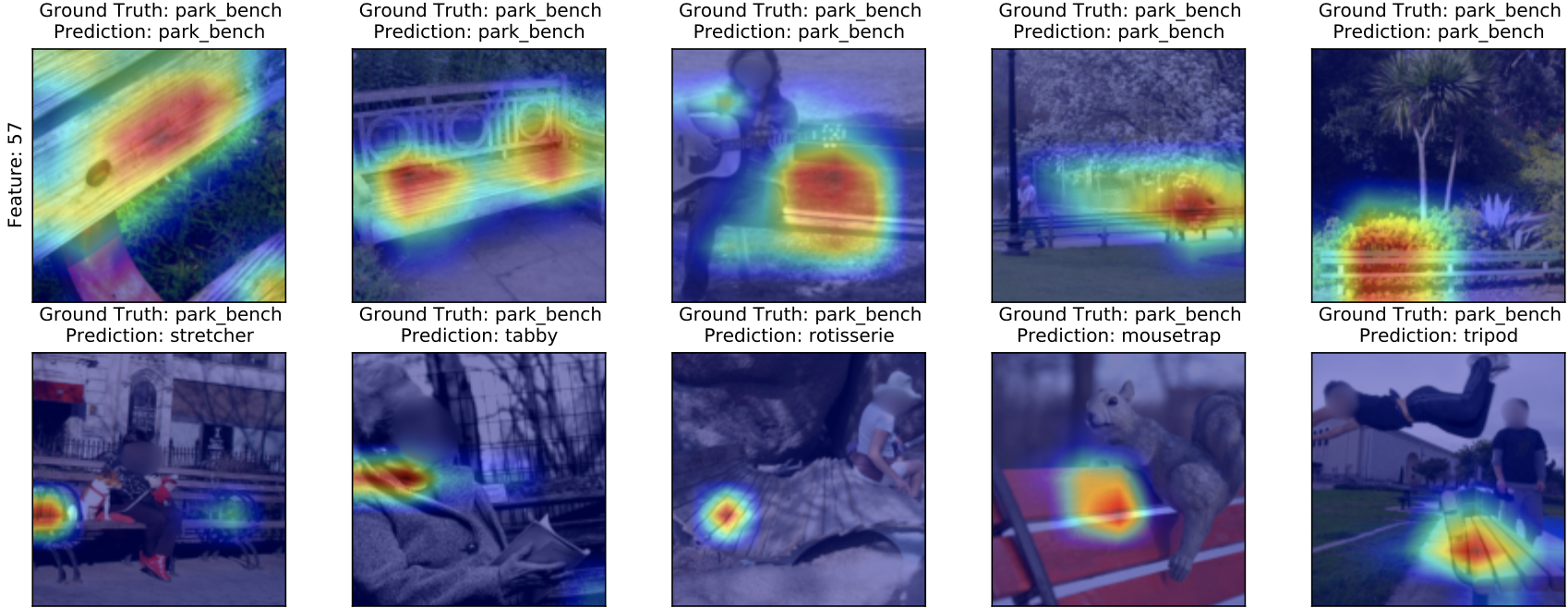}}

    \subfigure{\includegraphics[width = 0.8\textwidth, trim={0cm 0cm 0cm
0cm}, clip]{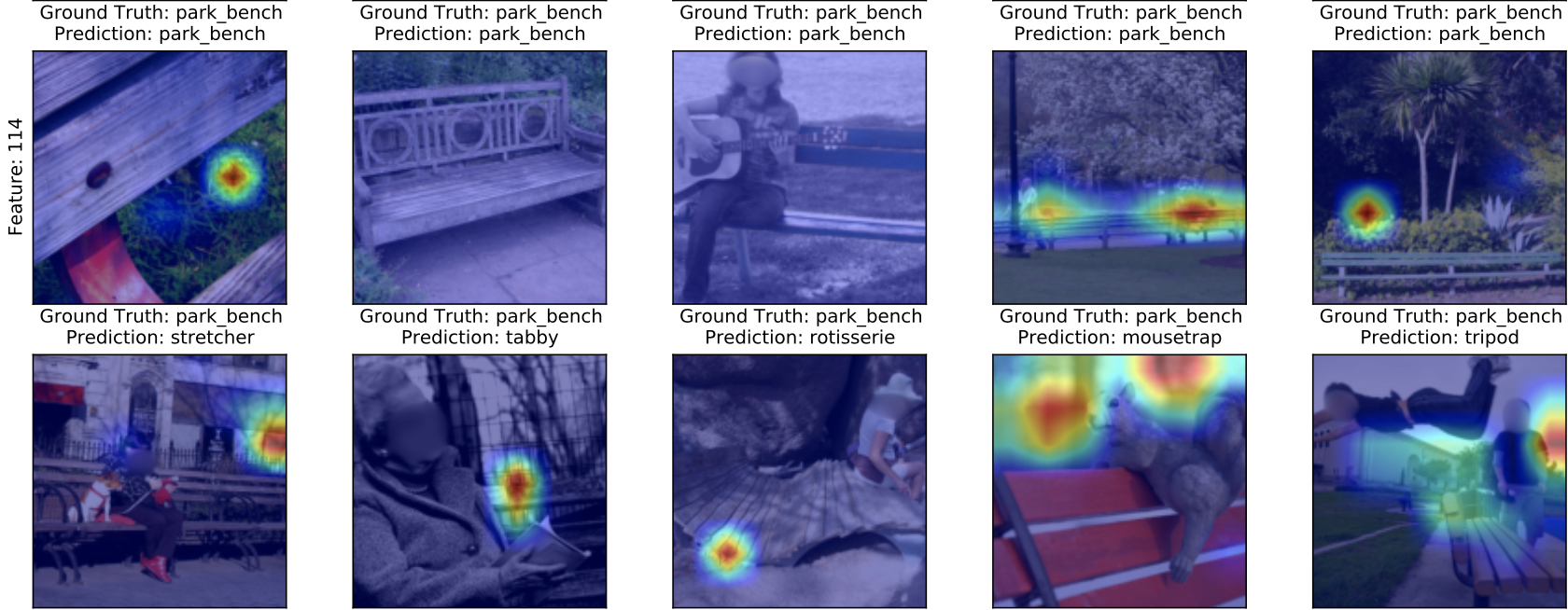}}
    
    \caption{\textbf{Heatmaps of discriminative and lowly activating features of SimCLR (Class - Park Bench)}: We plot the gradient heat maps of the top activating discriminative feature (by magnitude) for the given class and a lowly activating feature of the same class. We observe that discriminative features are more correlated with ground truth labels compared to lowly activating features in both correct and incorrect classification. The discriminative feature in correct classifications correspond to a unique physical attribute that may not exist (or be obfuscated) in mis-classified images.}
    \label{fig:more_grad_cams_park_bench}
\end{figure*}

\begin{figure*}
    \centering
    \subfigure{\includegraphics[width = 0.8\textwidth, trim={0cm 0cm 0cm
0cm}, clip]{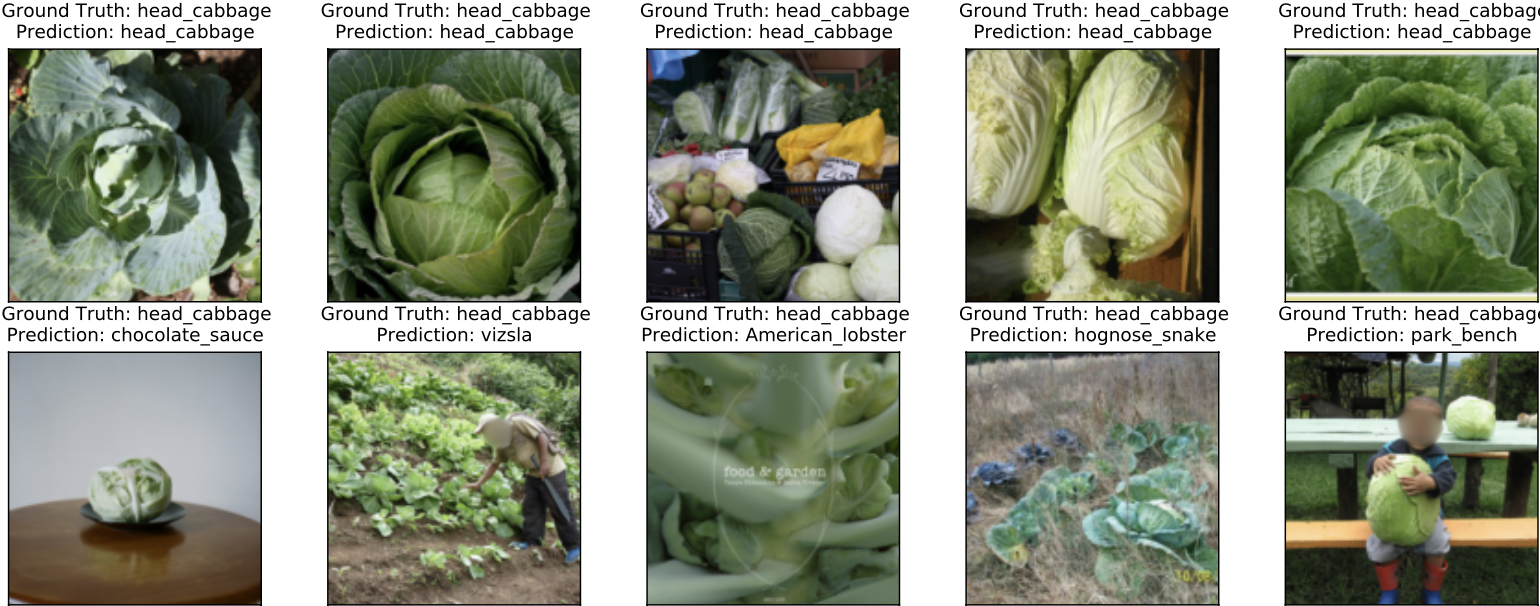}}
    \subfigure{\includegraphics[width = 0.8\textwidth, trim={0cm 0cm 0cm
0cm}, clip]{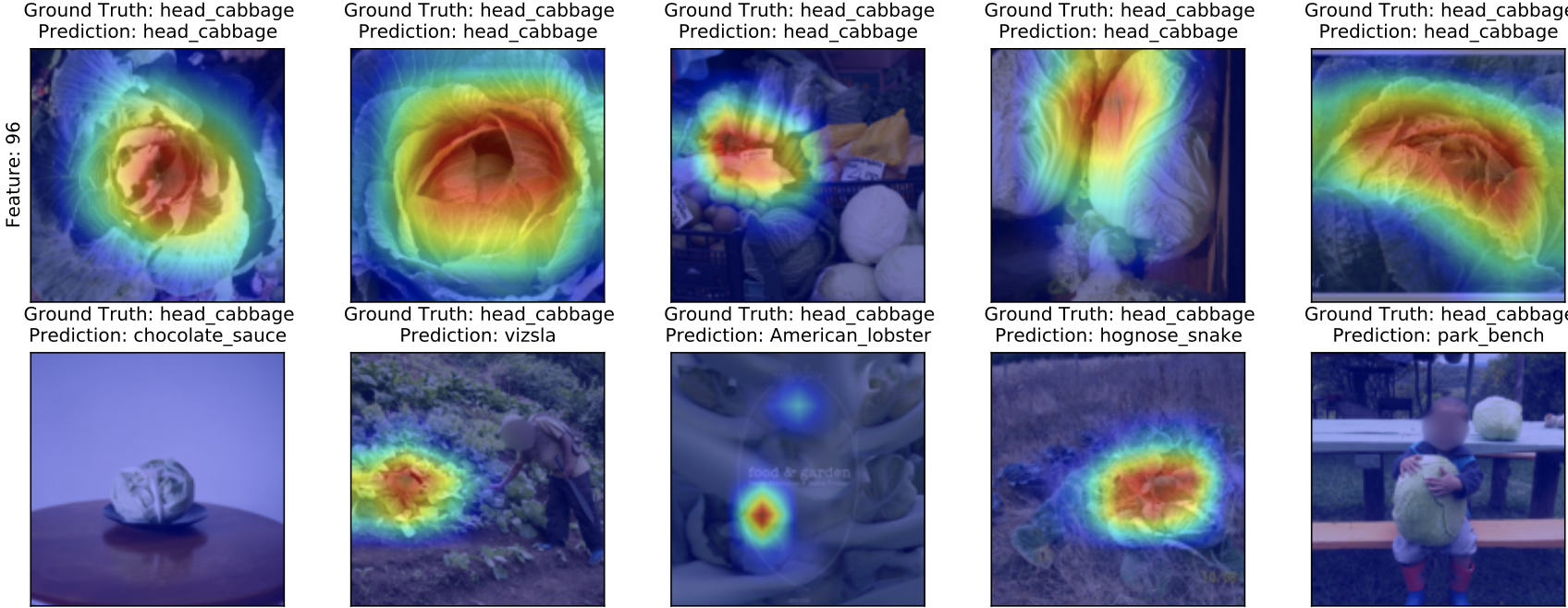}}
    \subfigure{\includegraphics[width = 0.8\textwidth, trim={0cm 0cm 0cm
0cm}, clip]{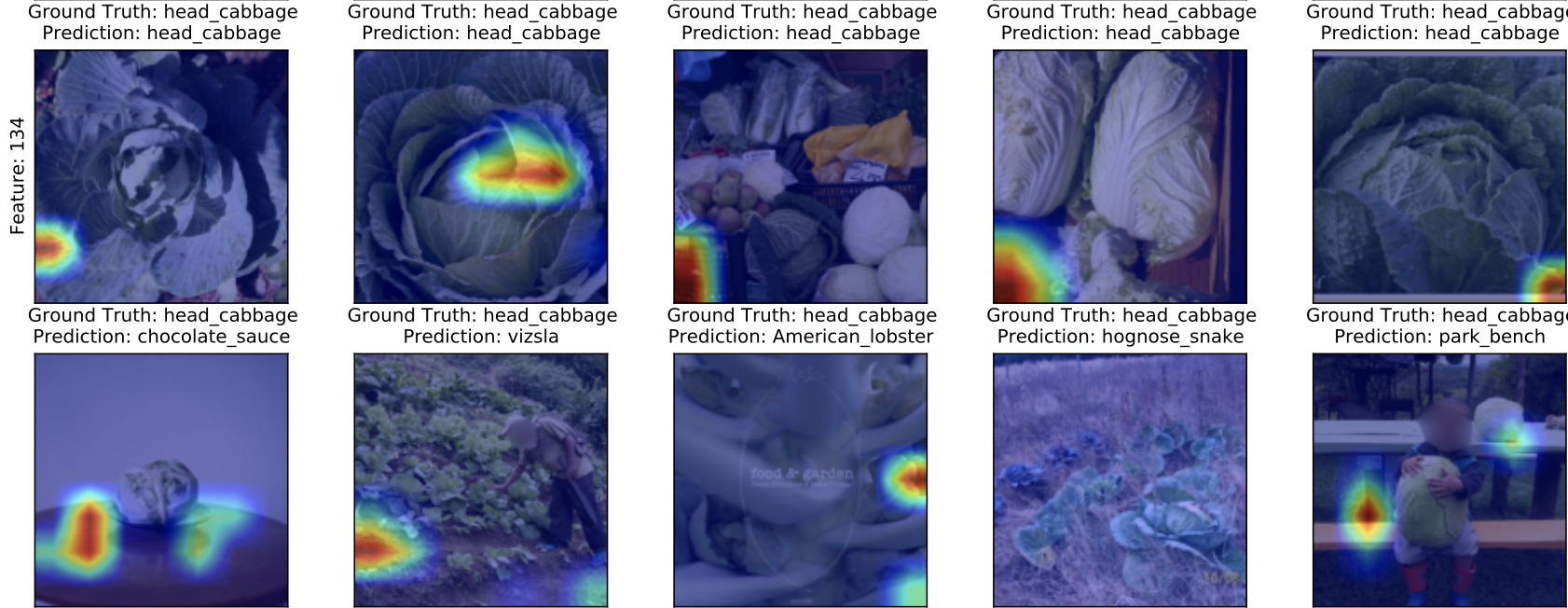}}
    
    \caption{\textbf{Heatmaps of discriminative and lowly activating features of SimCLR (Class - Head Cabbage)}: We plot the gradient heat maps of the top activating discriminative feature (by magnitude) for the given class and a lowly activating feature of the same class. We observe that discriminative features are more correlated with ground truth labels compared to lowly activating features in both correct and incorrect classification. The discriminative feature in correct classifications correspond to a unique physical attribute that may not exist (or be obfuscated) in mis-classified images.}
    \label{fig:more_grad_cams_head_cabbage}
\end{figure*}

\begin{figure*}
    \centering
    \subfigure{\includegraphics[width = 0.8\textwidth, trim={0cm 0cm 0cm
0cm}, clip]{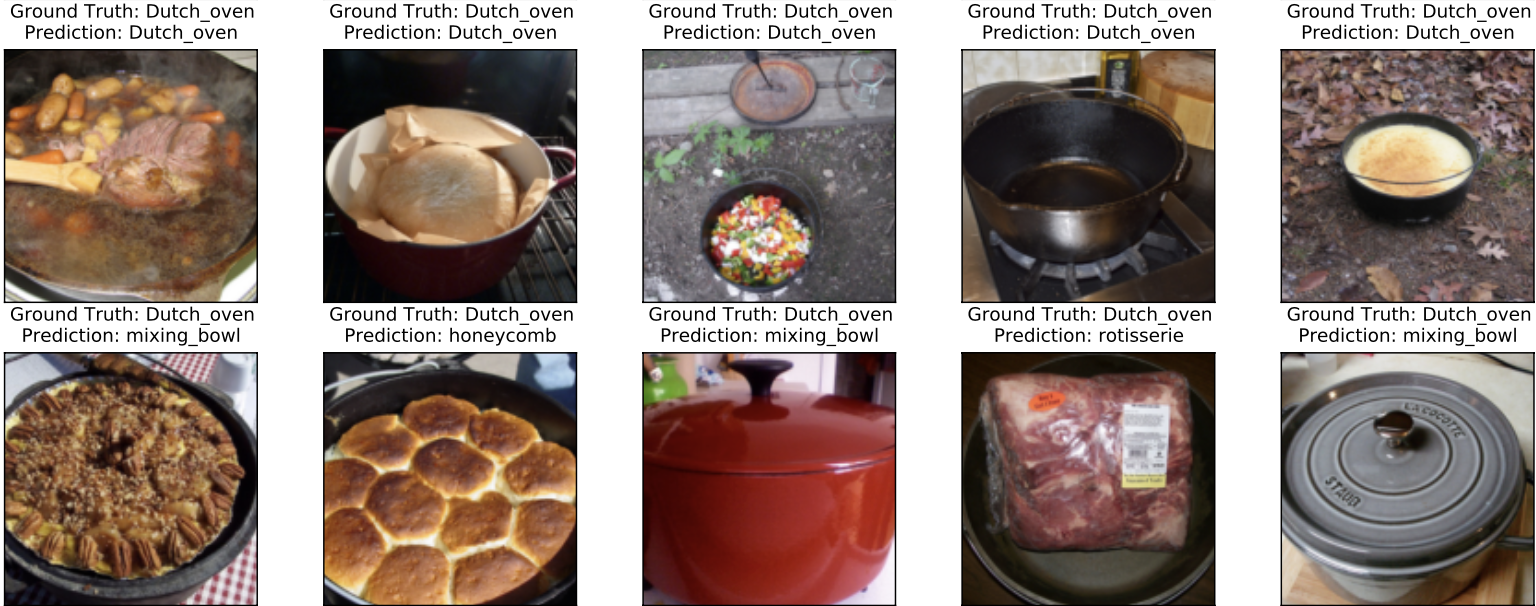}}

    \subfigure{\includegraphics[width = 0.8\textwidth, trim={0cm 0cm 0cm
0cm}, clip]{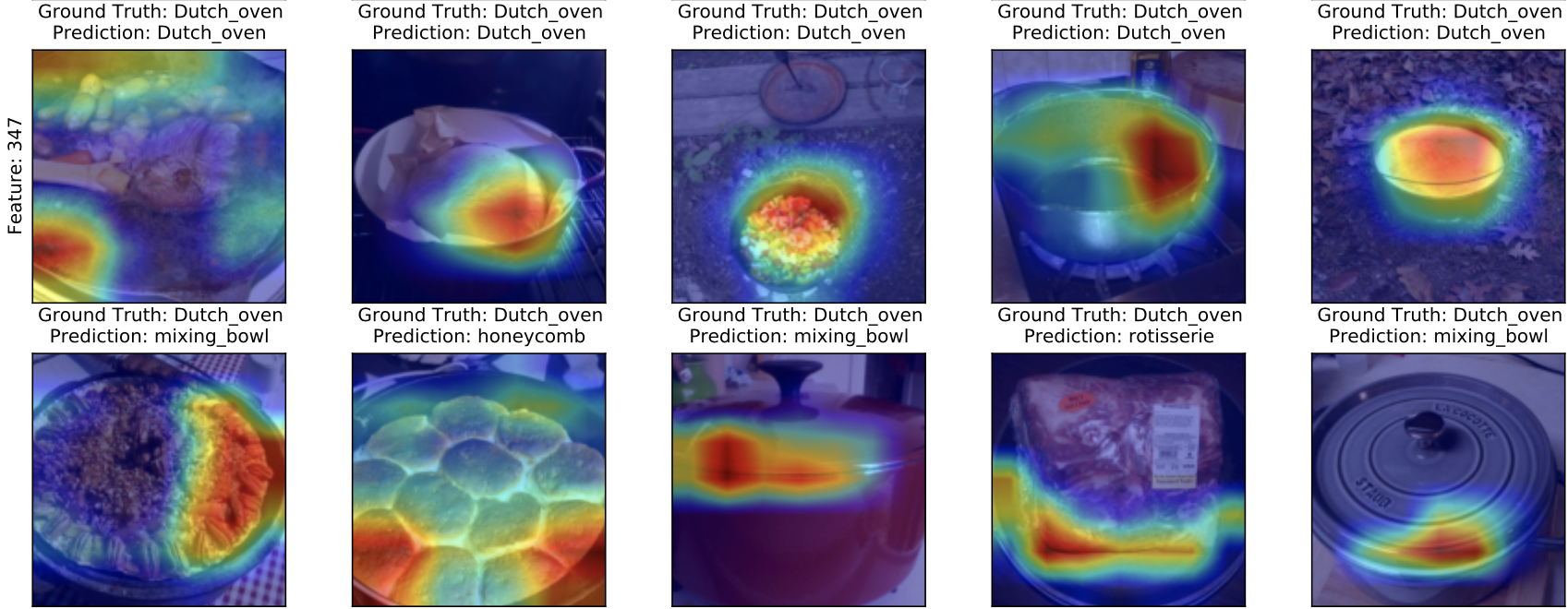}}

    \subfigure{\includegraphics[width = 0.8\textwidth, trim={0cm 0cm 0cm
0cm}, clip]{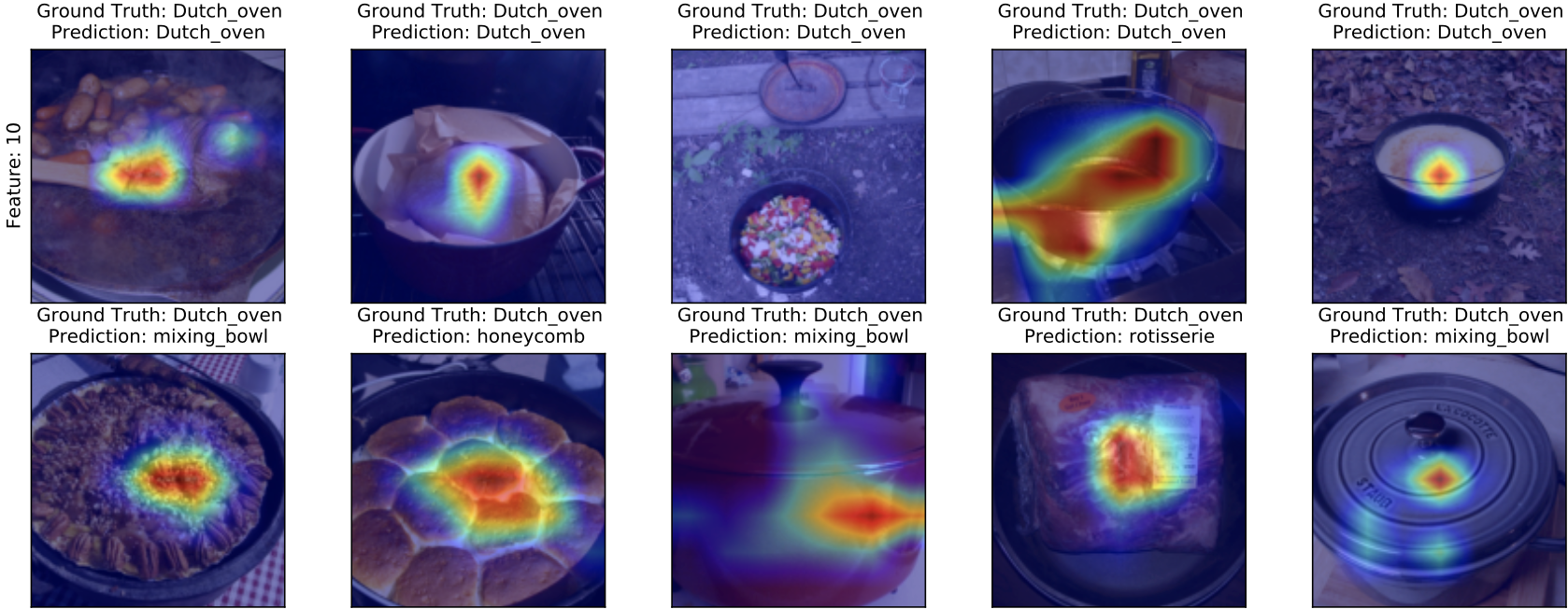}}
    
    \caption{\textbf{Heatmaps of discriminative and lowly activating features of SimCLR (Class - Dutch Oven)}: We plot the gradient heat maps of the top activating discriminative feature (by magnitude) for the given class and a lowly activating feature of the same class. We observe that discriminative features are more correlated with ground truth labels compared to lowly activating features in both correct and incorrect classification. The discriminative feature in correct classifications correspond to a unique physical attribute that may not exist (or be obfuscated) in mis-classified images.}
    \label{fig:more_grad_cams_dutch_oven}
\end{figure*}